\documentclass[11pt]{article}
\pdfoutput=1
\usepackage{etoolbox}
\newtoggle{iclr}
\togglefalse{iclr}

\usepackage[T1]{fontenc}
\usepackage[utf8]{inputenc}
\usepackage{times}
\usepackage[letterpaper,margin=1in]{geometry}
\usepackage{microtype}

\usepackage{amsmath,amssymb,amsthm,mathtools}
\usepackage{bm}
\usepackage{scalerel}
\usepackage[authoryear,square]{natbib}

\usepackage{color-edits}
\usepackage{enumitem}
\usepackage{xcolor}
\definecolor{mainlinkcolor}{HTML}{004C4C}
\colorlet{maintheoremcolor}{black}
\usepackage{graphicx}
\setkeys{Gin}{height=.85\textheight,keepaspectratio}
\usepackage[font=small,labelfont=bf]{caption}
\usepackage{tcolorbox}
\tcbuselibrary{skins,breakable}
\usepackage[breaklinks=true,colorlinks=true,citecolor=mainlinkcolor,
  linkcolor=mainlinkcolor,urlcolor=mainlinkcolor]{hyperref}
\usepackage{thm-restate}
\usepackage[capitalise,nameinlink,noabbrev]{cleveref}

\numberwithin{equation}{section}

\newtheoremstyle{mainplain}{6pt}{6pt}{\itshape}{}
  {\bfseries\color{maintheoremcolor}}{.}{.5em}{}
\newtheoremstyle{maindefinition}{6pt}{6pt}{\normalfont}{}
  {\bfseries\color{maintheoremcolor}}{.}{.5em}{}
\newtheoremstyle{mainremark}{6pt}{6pt}{\normalfont}{}
  {\itshape\color{maintheoremcolor}}{.}{.5em}{}
\theoremstyle{mainplain}
\newtheorem{theorem}{Theorem}[section]
\newtheorem{proposition}[theorem]{Proposition}

\theoremstyle{maindefinition}
\newtheorem{definition}[theorem]{Definition}

\theoremstyle{mainremark}

\newtcolorbox{maintheorembox}{enhanced jigsaw,breakable,
  colback=blue!60!gray!5,colframe=gray!50,boxrule=0pt,arc=2mm,
  boxsep=0pt,left=6pt,right=6pt,top=4pt,bottom=4pt,
  before skip=8pt,after skip=8pt}
\addtotheorempreheadhook{\begin{maintheorembox}}
\addtotheorempostfoothook{\end{maintheorembox}}

\allowdisplaybreaks

\newcommand{\R}{\mathbb{R}}

\newcommand{\Normal}{\mathcal{N}}

\title{Understanding Mode Collapse in Diffusion Models under Classifier-Free Guidance}
\author{%
  Daniel Pfrommer\textsuperscript{1}, Chenyang Yuan\textsuperscript{2}, Frank Permenter\textsuperscript{2}, Ali Jadbabaie\textsuperscript{1}\\[0.5em]
  {\small \textsuperscript{1}Massachusetts Institute of Technology\quad
    \textsuperscript{2}Toyota Research Institute}\\[0.3em]
  {\footnotesize
    \textsuperscript{1}\href{mailto:dpfrom@mit.edu}{\texttt{\{dpfrom, jadbabai\}@mit.edu}},
    }\\
  {\footnotesize
    \textsuperscript{2}\href{mailto:chenyang.yuan@tri.global}{\texttt{\{chenyang.yuan,frank.permenter\}@tri.global}}}
}
\date{November, 2026}

\begin{document}

\maketitle

\begin{abstract}
  The use of classifier-free guidance in diffusion and flow models has proven critical to their empirical success, as it allows for the controllable generation of lower-temperature and higher-fidelity samples. Despite its widespread use, classifier-free guidance remains both empirically and theoretically poorly understood, owing to the difficulty of solving the associated transport equations for arbitrary normalizing flows. In this paper, we observe that the effects of classifier-free guidance arise from a mismatch between the representation capabilities of the conditional and unconditional distributions, wherein the conditional distribution exhibits higher complexity and fidelity. We propose an analytical model to capture this observation: the conditional distribution is a Gaussian mixture whose means lie on an ellipsoid, and the unconditional distribution is a single Gaussian approximating the mixture. The flows arising from this mixture are closed under scaling, enabling us to derive analytic closed-form solutions for the guided distribution at all noise scales and guidance levels. Our analytic solution shows that classifier-free guidance induces (a) a shift along a mean-axis direction and (b) a persistent skew toward, and concentration around, a singular point corresponding to the conditional sample starting from zero noise. We validate our theoretical predictions on pretrained, large-scale image diffusion models.

\end{abstract}

%!TEX root = ../main.tex

\section{Introduction}
\label{sec:introduction}

Over the last several years, diffusion models have become the dominant paradigm for generative modeling across a wide spectrum of continuous data modalities, ranging from image \citep{zheng2026diffusion}, audio, and video generation \citep{yang2025cogvideox} tasks to computational biology \citep{abramson2024accurate}, robotics \citep{wang2022diffusion}, and time-series forecasting problems \citep{ding2026simdiff}. The fundamental principle that unifies diffusion-style generation methods is the construction of a normalizing flow that smoothly interpolates between data samples and random noise. The resulting flow can easily be learned by minimizing a standard supervised loss and sampled using either stochastic (SDE-based) \citep{ho2020denoising} or deterministic (ODE-based) \citep{song2020denoising} sampling methodologies.

A crucial component in the success of diffusion across these domains is the ability to amplify the performance of a conditional generative model via the construction of \emph{guided} flows \citep{ho2022classifier}, which linearly combine the learned flow of the unconditional and conditional distributions at inference time to enhance the features of the conditional distribution. The adoption of guidance extends even beyond diffusion models, as various other generative paradigms have reported performance improvements by including guidance-style contrastive terms in their formulation~\citep{sanchez2023stay,chang2023muse,nisonoff2025unlocking}.

While guided flows generally produce both lower-temperature and higher-quality samples than either their unconditional or conditional counterparts, they break the fundamental data-to-noise interpolative foundation on which the theory of diffusion models is built \citep{du2023reduce}. As a result, the effects of guidance on the resulting sample distribution remain comparatively poorly understood. Previous work has identified the ability of guidance to selectively amplify or suppress specific feature directions~\citep{li2026towards} and suggests guided flows weakly approximate exponentially-weighted density products of the underlying flows~\citep{liu2022compositional,du2023reduce,bradley2024classifier}. The derivation of exact closed-form solutions has previously necessitated either (1) restricting the analysis to Gaussian distributions, for which the underlying flows are purely linear~\citep{wu2024theoretical,li2026towards}, or (2) asymptotic analysis or integration of a correction term over the associated sample trajectories~\citep{zheng2023characteristic,skreta2025feynman,jiang2026analytic,pavasovic2025classifier}. For an extended discussion of the prior work, see \Cref{app:prior_work}.

\textbf{Contributions.} In this work, we make several key advancements in the understanding of classifier free guidance from both theoretical and empirical perspectives:
\iftoggle{iclr}{\begin{enumerate}[nosep,leftmargin=1.5em]}{\begin{enumerate}}
    \item We derive what, to our knowledge, is the first closed-form solution for a guided density involving a multi-modal conditional distribution, for which the underlying flow is nonlinear (\Cref{cor:one-dimensional}).
    \item We consider a model for guidance where the conditional distribution is a mixture, while the unconditional distribution is represented by a single Gaussian. We argue that this better represents classifier-free guidance, as in practice unconditionally-generated samples exhibit poor fidelity, suggesting that unconditional flows are less expressive. Our setting also captures techniques such as self-guidance \citep{li2025self} and auto-guidance \citep{karras2024guiding}, where a lower-capacity conditional model is used in place of the unconditional flow.
    \item Our analysis (\Cref{thm:guided-density}) exactly predicts both (1) the simultaneous magnification of features across different, component-specific axes and (2) concentration around a single point represented by setting the initial noise to zero. We additionally show that this mode-collapse effect is \emph{bounded} and derive a limit on the concentration around the collapsed mode.
\end{enumerate}
We empirically validate our findings on pretrained, large-scale image diffusion models for text- and class-conditioned settings. We show the prevalence of both the hypothesized cluster-specific feature magnification effect, in addition to a limited degree of modality collapse to the mode represented by the zero-initialized sample, in the specific manners predicted by our model.

%!TEX root = ../main.tex

\section{Diffusion Preliminaries}
\label{sec:interpolants}
\begin{figure}[!t]
  \centering
  \includegraphics[width=0.9\linewidth]{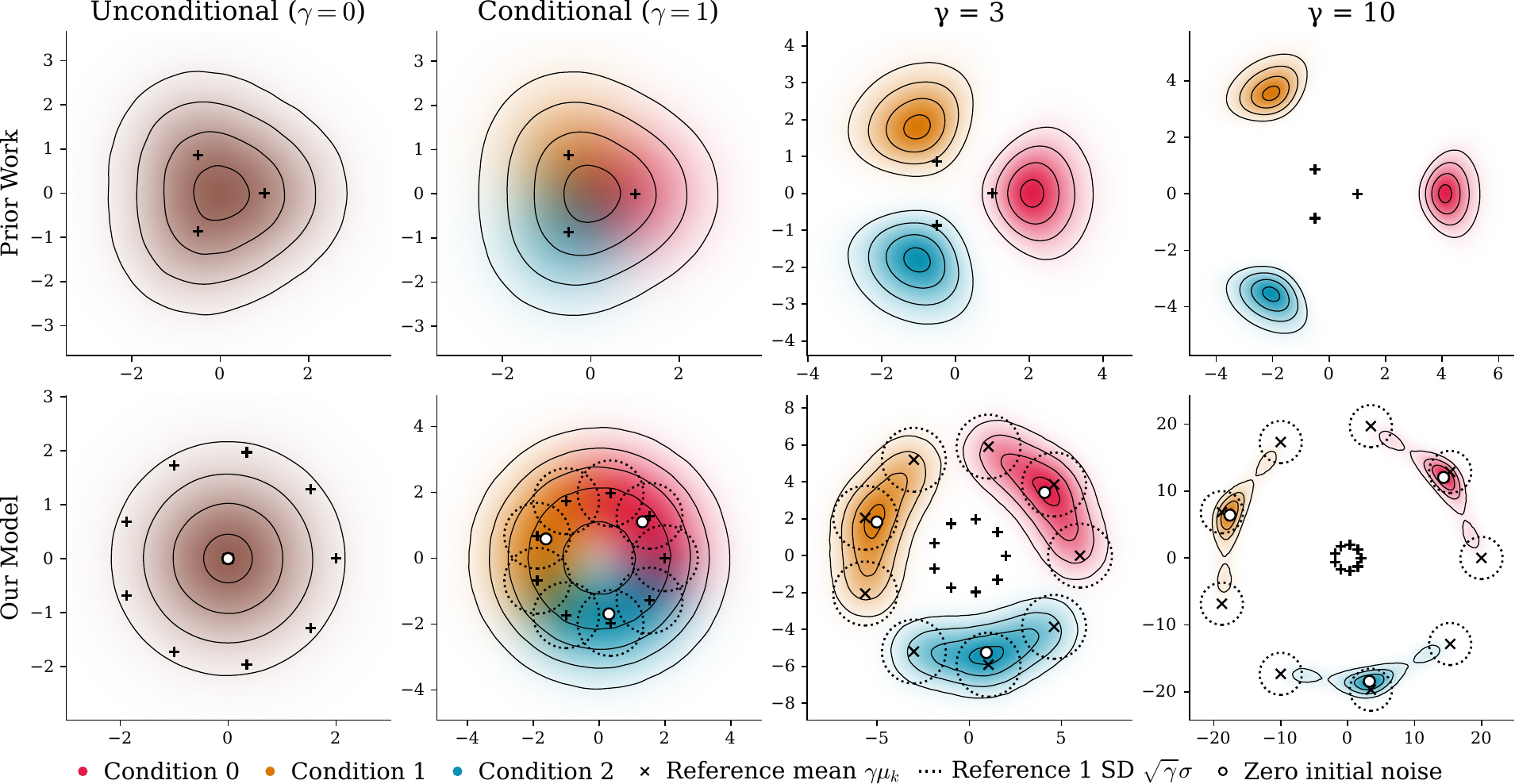}
  \caption{\textbf{Top:} A standard model for classifier-free guidance \citep{ho2022classifier}, where the conditional distribution is a single component of the full unconditional mixture distribution. \textbf{Bottom:} Our model, where each conditional distribution is a Gaussian mixture (pictured here with three components per condition) and the unconditional is a single Gaussian approximating the marginalized mixture. Our model admits an analytical closed-form solution, capturing the \emph{feature-magnification} and \emph{modality collapse} effects outlined in \Cref{thm:guided-density}.}
  \label{fig:toy-2d}
\end{figure}

\newcommand{\rmd}{\mathrm{d}}
\newcommand{\Exp}{\mathbb{E}}

We base our formulation of diffusion on the stochastic interpolant framework of
\citet{albergo2022building,albergo2025stochastic}. In this framework, we construct a random variable $X_t$ that interpolates between samples from a data distribution $X_0 \sim p(x)$ and pure noise $X_1$. We consider the standard linear interpolation given by a \emph{diffusion schedule} $\alpha(t), \sigma(t)$ weighting the data and noise respectively, for which the interpolation is equivalent to the standard diffusion forward process used in \cite{ho2020denoising, lipman2022flow, zheng2023guided, song2020denoising} and related literature.

\textbf{Notation:} For $A\succ0$, we use $\|\cdot\|_{A}$ to denote the Mahalanobis norm $\|z\|_A:=\sqrt{z^\top Az}$. We use uppercase ($X$) to denote matrices or random variables and lowercase ($x$) for vectors or scalars.

\begin{definition}[Noising Process]
\label{def:interpolant}
Let $X_0$ be a continuous random variable supported on $\R^n$ with density $p_0(x)$ and finite second moment, and let
$X_1\sim\Normal(0,\Sigma)$ be independent of $X_0$, where
$\Sigma\in\R^{n\times n}$ is symmetric positive definite. For continuously
differentiable functions $\alpha,\sigma\colon[0,1]\to\R$, define,
\begin{equation}
  X_t:=\alpha(t)X_0+\sigma(t)X_1,
  \label{eq:interpolant}
\end{equation}
where the noise schedule $(\alpha, \sigma)$ satisfies,
\begin{equation}
  \alpha(0)=1,\quad \sigma(0)=0,
  \qquad \alpha(1)=0,\quad \sigma(1)=1, \qquad \alpha(t)^2+\sigma(t)^2=1,
\end{equation}
with $\alpha$ non-increasing and $\sigma$ non-decreasing. For simplicity we only consider variance-preserving interpolants such that $\alpha(t)^2 + \sigma(t)^2 = 1$ for all $t$.
\end{definition}

The terminal distribution $X_1$ is generally taken to be $\Normal(0, I)$, i.e. $\Sigma = I$. We leave $\Sigma$ generic in our formulation as the added flexibility simplifies the proof of our main result, \Cref{thm:guided-density}, which relies on the construction of a flow with a misspecified $p_1(x)$ distribution.

\begin{restatable}[Transport equation, \citealp{albergo2025stochastic}]{proposition}{transportequation}
The densities $p_t(x)$ associated with $X_t$ as defined in \Cref{def:interpolant} satisfy the continuity equation,
\begin{equation}
  \partial_t p_t+\nabla_x\cdot(v_t p_t)=0,
  \label{eq:transport}
\end{equation}
in the weak sense, where
\begin{equation}
  v_t(x)=\Exp[\dot\alpha(t)X_0+\dot\sigma(t)X_1\mid X_t=x].
  \label{eq:velocity}
\end{equation}
\end{restatable}
By using the stochastic reverse-time sampler we can traverse the marginals $p_t(x)$ from $t = 1$ to $t=0$ and thereby generate new samples from the data distribution $p_0(x)$.

\begin{proposition}[Reverse-time sampler, \citealp{albergo2025stochastic}]
Assume $p_t$ is positive and sufficiently smooth, and that the SDE below is well posed and its Fokker--Planck equation has a unique probability-density solution. Define $s_t(x)=\nabla_x\log p_t(x)$ and let
$w\colon[0,1]\to\R_{\geq0}$ be continuous. With $W_\ell$ a standard
$n$-dimensional Brownian motion, set
\begin{align}
  \rmd X_\ell^{\mathrm R}
  =\bigl[-v_{1-\ell}(X_\ell^{\mathrm R})
       +w(\ell)s_{1-\ell}(X_\ell^{\mathrm R})\bigr]\rmd\ell
       +\sqrt{2w(\ell)}\,\rmd W_\ell,  \quad
  X_0^{\mathrm R}\sim\Normal(0,\Sigma).\label{eq:reverse-sde}
\end{align}
Then $X_\ell^{\mathrm R}$ has density $p_{1-\ell}$.
\end{proposition}
We consider the special case of $w=0$, where \eqref{eq:reverse-sde} simplifies to the ODE,
\begin{equation}
  \dot X_t=v_t(X_t),\qquad X_1\sim\Normal(0,\Sigma),
  \label{eq:sampling-ode}
\end{equation}
integrated from $t=1$ toward $t=0$.
For $m_t(x) :=\Exp[X_0\mid X_t=x]$ and
$e_t(x) :=\Exp[X_1\mid X_t=x]$, the discretized interpolant update from
$t$ to $s<t$ is precisely noiseless DDIM~\citep{song2020denoising}:
\begin{align}
  x_s&=\alpha(s)m_t(x_t)+\sigma(s)e_t(x_t) \label{eq:ddim-update}\\
     &=x_t+[\alpha(s)-\alpha(t)]m_t(x_t)
           +[\sigma(s)-\sigma(t)]e_t(x_t).\nonumber
\end{align}

\paragraph{Conditional sampling and classifier-free guidance.}
For a conditioning variable $C$, assume the noise $X_1$ is independent of
$(X_0,C)$. The conditional velocity is
\begin{equation}
  v_t(x\mid c)=\Exp[\dot\alpha(t)X_0+\dot\sigma(t)X_1\mid X_t=x,C=c].
\end{equation}
The exact conditional flow $v_t(x \mid c)$ transports the conditional data distribution. However, in practice, classifier-free guidance is used to trade sample diversity for fidelity~\citep{ho2022classifier}, with guidance strength $\gamma$.
\begin{definition}[Classifier-Free Guidance]
\label{def:guidance}
Given conditional flow $v_t(\cdot | c)$ and unconditional flow $v_t(\cdot)$, define, for
guidance strength $\gamma\in\R$ the guided flow,
\begin{equation}
  v_{\gamma,t}(x|c)=\gamma v_t(x|c)+(1-\gamma)v_t(x).
  \label{eq:guidance-definition}
\end{equation}
The choice $\gamma=1$ recovers the original conditional flow, while $\gamma>1$
extrapolates beyond it.
\end{definition}

Prior work on guidance in toy low-dimensional settings from which practitioners have drawn inspiration (see e.g.  Figure 3.1 of \cite{zheng2023guided}, reproduced here as the top row in \Cref{fig:toy-2d}) assume that the unconditional distribution consists of the full data mixture while conditioning isolates a subset of features, i.e. the unconditional distribution $p(x)$ is given by $\sum_{k=1}^K \pi_k \Normal(\mu_k, \Sigma_k)$ for some parameters $\pi_k, \mu_k, \Sigma_k$ while the conditional distribution $p(x|c)$ is given by $\Normal(\mu_{j_c}, \Sigma_{j_c})$ for some $j_c$.

\paragraph{Motivation for our analytical model.} Assuming the unconditional distribution is a more complex mixture would mean that the unconditional model is more expressive than the conditional model. As previously discussed, we believe this is not reflective of classifier-free guidance in practice, as (1) unconditional models exhibit far worse fidelity than their conditional counterparts \citep{ho2022classifier} and (2) our model captures techniques such as self-guidance \citep{li2025self} and autoguidance \citep{karras2024guiding}, which use less-expressive conditional models in place of the ``unconditional'' flow. We follow \cite{jin2026stage} in assuming that the unconditional flow is a ``coarse'' approximation and can be represented by a single Gaussian, while using a mixture to represent the conditional distribution. \Cref{fig:imagenet-tsne} demonstrates precisely this structure via t-SNE embeddings of DINO-V2 features of generated ImageNet samples while \Cref{fig:toy-2d} contrasts our setup with that of prior work and highlights the magnification and mode-collapse effects that our exact form predicts.

% The remainder of this paper is structured as follows. In \Cref{sec:exact-density} we formally present the special class of Gaussian mixtures with common radius means for which we derive our theoretical results and closed form. We show in \Cref{sec:experiments} that the expansion and tail effects we predict can be directly observed in real image models. Lastly, in \Cref{sec:conclusion} we discuss the limitations of our approach and potential extensions.

\begin{figure}[t]
  \centering
  \includegraphics[width=0.9\linewidth]{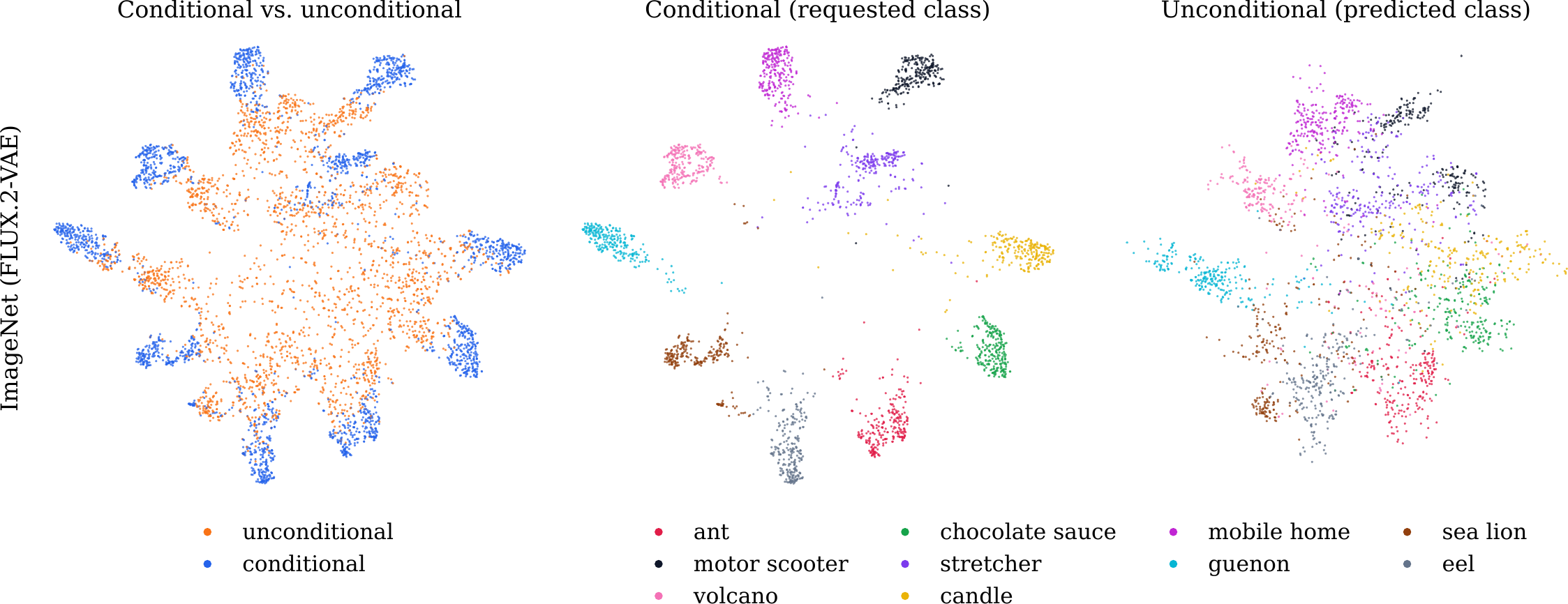}
  \caption{ Empirical evidence suggests that the conditional ($\gamma=1$) generation exhibits more distinctly class-specific samples of higher complexity relative to unconditional generation ($\gamma=0$). We visualize t-SNE embeddings of DINO-V2 features of ImageNet samples using conditional vs.~unconditional diffusion models for 10 different classes. Conditional samples are generated using class conditioning whereas unconditional samples are generated using rejection sampling with a ResNet-based classifier. See \Cref{fig:image-tsne-sets} for more comparisons with different classes. }
  \label{fig:imagenet-tsne}
\end{figure}

\section{Analytical Solutions to Guidance with Mixture-Based Conditional Distributions}
\label{sec:exact-density}

We now present our main theoretical model and results, with proofs for this section deferred to \Cref{app:proofs}. Concretely, we model the unconditional distribution $p_0(x)$ by $\Normal(x;\bar\mu, \Sigma)$ and the conditional distribution $p_0(x|c)$ by a mixture of Gaussians which lie on a common-radius ellipse around $\bar\mu$. Throughout our analysis, we consider a fixed conditioning value $c$ and write $p(x|c)$ to distinguish the conditional $\gamma = 1$ distribution from $p_0(x)$, the unconditional $\gamma = 0$ distribution. 

\begin{definition}[Common-radius Gaussian Mixture]
\label{def:crgm}
A common-radius Gaussian mixture is a random variable $X$ with density $q(x)$ such that
\begin{equation}
  q(x)=\sum_{k=1}^K\pi_k\Normal(x;\mu_k,\Sigma),
  \qquad \pi_k>0,\qquad \sum_{k=1}^K\pi_k=1, \label{eq:crgm}
\end{equation}
where $\Sigma\succ0$ and there exists some center $\bar\mu\in\R^n$ and $R\geq0$ such that
$\|\mu_k-\bar\mu\|_{\Sigma^{-1}}=R$ for every $k$. We say that a conditional distribution $q(x|c)$ satisfies \eqref{eq:crgm} at $c$ if $q(\cdot |c)$ matches the above density.
\end{definition}

\begin{definition}[Common-radius Conditioning Property]\label{assum:main}We say that for a fixed $C = c$, the densities $p_0(x|c)$ and $p_0(x)$ in $X$ satisfy a common-radius conditioning property if, as a function of $x$, $p_0(x|c)$ has the form of \Cref{eq:crgm} for some $\mu_1, \ldots, \mu_k, \bar\mu,$ and $\Sigma$ and $p_0(x) = \mathcal{N}(x;\bar\mu, \Sigma)$.
\end{definition}

Under \Cref{assum:main}, the flow $v_t(x)$ for $p_0(x) = \mathcal{N}(x;\bar\mu, \Sigma)$ is simply $v_t(x)=\dot\alpha(t)\bar\mu$ (see \Cref{prop:mixture-interpolant}), while the conditional vector field $v_t(x|c)$ has the following special form:
 
\begin{restatable}[Mixture velocity]{proposition}{mahalanobismixtureflow}
\label{prop:mahalanobis-mixture-flow}
Suppose that $X_0 \,|\, C=c$ is a common-radius mixture as in \Cref{def:crgm} with center $\bar\mu$, means $\mu_1, \ldots, \mu_K$, and covariance $\Sigma$. Let $X_1 \sim \mathcal{N}(0,\Sigma)$ be independent of $X_0 \,|\, C=c$. Then the velocity $v_t(x | c)$ defined in \Cref{def:interpolant} satisfies,
\begin{equation}
  \begin{aligned}
  v_t(x | c)&=\dot\alpha(t)\sum_k w_k(t,x)\mu_k,\\
  w_k(t,x)&=\frac{\pi_k e^{\alpha(t)(x-\alpha(t)\bar\mu)^\top\Sigma^{-1}(\mu_k-\bar\mu)}}
                   {\sum_j\pi_j e^{\alpha(t)(x-\alpha(t)\bar\mu)^\top\Sigma^{-1}(\mu_j-\bar\mu)}}.
  \end{aligned}
  \label{eq:common-radius-velocity}
\end{equation}
\end{restatable}
\textbf{Approach Sketch:} The key property of \Cref{prop:mahalanobis-mixture-flow} which enables our analysis is that the family of flows arising from \Cref{def:crgm} is \emph{closed} under scaling when $\bar\mu = 0$. Multiplying such a $v_t(x|c)$ by $\gamma$ with $\bar\mu = 0$ results in correspondingly scaled means $\gamma \mu_k$ and covariances $\gamma \Sigma$. This can be directly extended to the setting where $\bar\mu \neq 0$ by considering the $\mu_k - \bar\mu$ offset instead. Given $\mu_1, \ldots, \mu_k$, $\bar\mu$, and $\Sigma$, we define the \emph{auxiliary density} $\hat p_{\gamma, t}$,
\begin{equation}
  \hat p_{\gamma,t}(x | c):=\sum_{k=1}^K\pi_k
       \Normal\bigl(x;\alpha(t)[\bar\mu+\gamma(\mu_k-\bar\mu)],\gamma\Sigma\bigr).
  \label{eq:aux_pdf}
\end{equation}
The above $\hat p_{\gamma, t}$ satisfies the continuity equation \eqref{eq:transport}
for $v_{\gamma,t}$ (i.e. $\partial_t \hat{p}_{\gamma, t} + \nabla_x \cdot (v_{\gamma,t}\hat{p}_{\gamma, t}) = 0$), but with a mismatched boundary condition. At $t = 1$,  $\hat p_{\gamma,1}(x)$ is the $\gamma$-dependent
$\Normal(0,\gamma\Sigma)$, instead of the desired $\Normal(0, \Sigma)$. Sampling using the standard reverse process \eqref{eq:sampling-ode} using the desired $p_1(x)=\Normal(x;0,\Sigma)$ therefore reweights $\hat{p}_{t, \gamma}$ in a manner that \Cref{thm:guided-density} makes explicit:

\begin{restatable}[Guided Density]{theorem}{guideddensity}
\label{thm:guided-density}
Fix some $C = c$ for which the conditional density $p_0(x | c)$ satisfies \Cref{def:crgm} for $\mu_1, \ldots, \mu_K$, $\bar\mu$, and $\Sigma$ and let $p_0(x) = \mathcal{N}(\bar{\mu}, \Sigma)$. For any $\gamma > 0$ the guided velocity conditioned at $c$ is given by,
\begin{align}
  v_{\gamma,t}(x|c)=\dot\alpha(t)\left[\bar\mu+
    \gamma\sum_k w_k(t,x)(\mu_k-\bar\mu)\right]. \label{eq:main_guided}
\end{align}
Define $y_{\gamma, t}$ as a function which transports samples from $\hat{p}_{\gamma, 1}$ to $\hat{p}_{\gamma, t}$  through $v_{\gamma, t}$ with corresponding inverse map $y_{\gamma, t}^{-1}$, i.e. 
\begin{equation}
  y_{\gamma, t}(x)=z_t, \qquad \mathrm{where} \qquad
  \dot z_s=v_{\gamma,s}(z_s|c),\qquad z_1=x.
  \label{eq:characteristic-map}
\end{equation}
Then the unique family of probability densities $p_{\gamma,t}$ satisfying both the transport equation for the guided velocity $v_{\gamma, t}$ in \eqref{eq:main_guided} and the boundary condition $p_{\gamma,1}=\Normal(0,\Sigma)$ is,
\begin{align}
  p_{\gamma,t}(x|c)
  &= \underbrace{\gamma^{n/2}\exp\!\left(-\frac{\gamma-1}{2\gamma}
        \|y_{\gamma, t}^{-1}(x)\|_{\Sigma^{-1}}^2\right)}_{W_{\gamma,t}(x)}
       \hat p_{\gamma,t}(x|c),\label{eq:guided-density}
\end{align}
where $\hat{p}_{\gamma, t}$ is as defined in \Cref{eq:aux_pdf}. For $\gamma>1$, since $y_{\gamma,t}$ is a diffeomorphism, the reweighting function $W_{\gamma, t}$ satisfies
\begin{equation}
  W_{\gamma,t}(y_{\gamma,t}(0))=\gamma^{n/2}=\max_x W_{\gamma,t}(x),
  \qquad
  \lim_{\|x\|_{\Sigma^{-1}}\to\infty}W_{\gamma,t}(x)=0.
  \label{eq:central-outer-weight}
\end{equation}
\end{restatable}

Note that both the reweighting functions $W_{\gamma,t}(x)$ and $y_{\gamma,t}(x)$ are dependent on $v_{\gamma,t}(x|c)$ and therefore also implicitly depend on the fixed choice of $c$, although we omit this for notational convenience.

\begin{figure}[!t]
  \centering
  \includegraphics[width=0.8\linewidth]{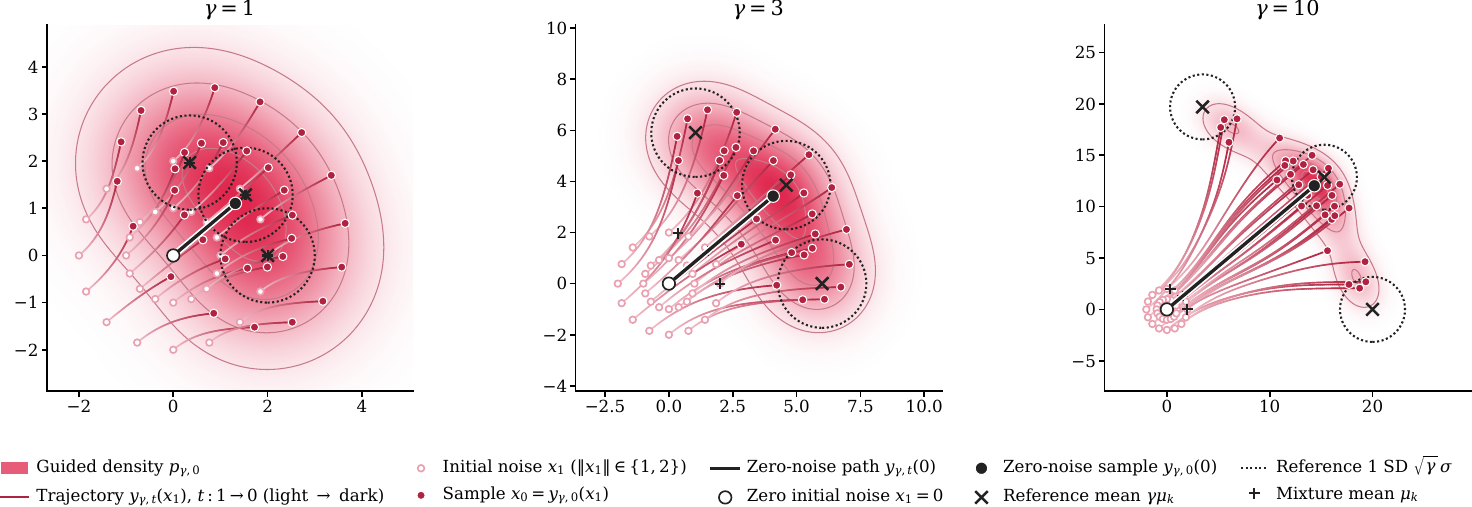}
  \caption{
  Guided flow map trajectories for a group of initial points while varying classifier-free guidance scales, derived from our analytical model in \cref{sec:exact-density}. Increasing guidance scale concentrates the guided distribution (shown as shaded density) towards the zero-noise sample $y_{\gamma,0}(0)$.}
  \label{fig:flow-map-traj}
\end{figure}

\iftoggle{iclr}{}{\newpage}

\textbf{Interpretation:} \Cref{eq:guided-density} highlights two complementary effects of guidance for $\gamma > 1$:
\iftoggle{iclr}{\begin{enumerate}[label=(\arabic*),nosep,leftmargin=1.5em]}{\begin{enumerate}[label=(\arabic*)]}
    \item \textit{Component-Specific Magnification}: The covariance $\Sigma$ and component-specific features $\mu_k - \bar\mu$ in $\hat{p}_{\gamma,t}$ are scaled by $\gamma$.
    \item \textit{Mode Collapse}: The resulting scaled distribution $\hat p_{t,\gamma}$ is then focused via $W_{t, \gamma}$ around $y_{\gamma, t}(0)$, the $x_t$ associated with $x_1 = 0$ under the guided flow. As a result of this effect, $p_{\gamma,t}$ is skewed towards $y_{\gamma, t}(0)$.
\end{enumerate}
These effects explain both (1) the exaggeration of class-specific features and (2) the loss of sample diversity under high guidance as $y_{\gamma,t}(0)$ is over-represented. Note that while (1) increases the data covariance, this effect is itself counteracted by the concentration (2), which depends on the specific placement of $\mu_1, \ldots, \mu_K$ and $\bar\mu$. In \Cref{sec:experiments} we will show that both of these effects can be visualized in image diffusion models by considering projections along the predicted directions.

Although \Cref{eq:guided-density} offers insights into the structure of guided distributions, computing $y_{\gamma, t}$  still requires solving an ODE. In the following section we show that for particular distributions the dependency on $y_{\gamma, t}$ may be removed either in part or in its entirety, yielding the first exact closed-form solution for a guided density with a nonlinear flow.

\subsection{Explicit Closed-Forms for the Guided Distribution}
\begin{figure}[!t]
  \centering
  \includegraphics[width=\linewidth]{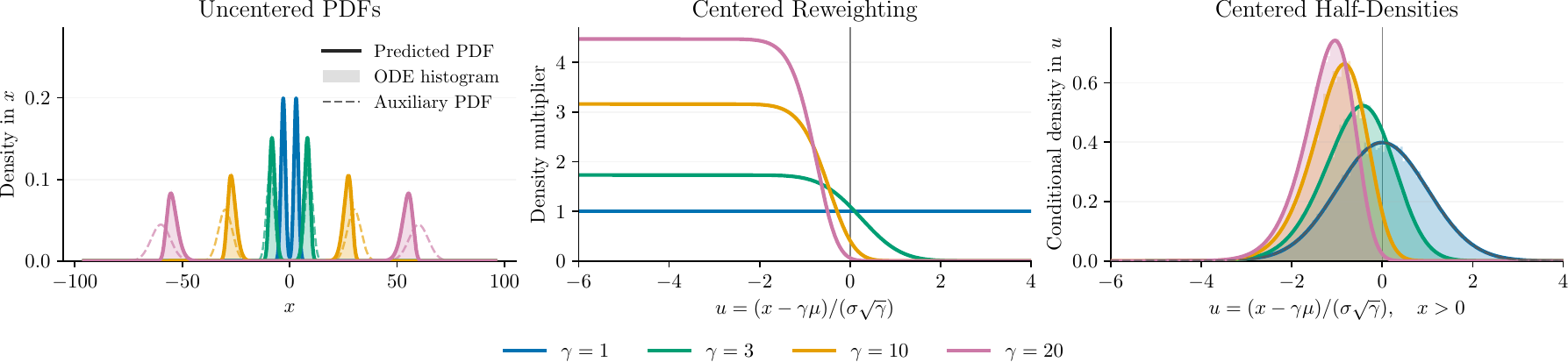}
  \caption{One-dimensional closed-forms for the $\gamma=1$ mixture $\tfrac12\Normal(-3,1)+\tfrac12\Normal(3,1)$ and $\gamma = 0$ unconditional $\Normal(0, 1)$, for $\gamma\in\{1,3,10,20\}$. From left to right:
  \textbf{(1)} the densities $\hat{p}_{\gamma, t}$ (dashed) and $p_{\gamma, t}$ (solid), \textbf{(2)} the density multiplier $W_{\gamma,0}$, in $(x - \gamma \mu)/(\sigma \sqrt{\gamma})$ coordinates, and \textbf{(3)} the positive-half densities of $p_{\gamma, 0}$, also in rescaled coordinates. Lines use the exact forms in \Cref{cor:one-dimensional} while shaded areas use a histogram with 100,000 integrated samples.}
  \label{fig:reweighting-1d}
\end{figure}

In one dimension the CDF of the auxiliary distribution $\hat{p}_{t,\gamma}$ is constant on $y_{\gamma, t}(x)$ for all $t$, which
allows us to eliminate $y_{\gamma,t}^{-1}(x)$ from \Cref{eq:guided-density} and write $p_{\gamma,t}(x|c)$ with $W_{\gamma,t}$ expressed in terms of the standard normal CDF $\Phi$ and its inverse $\Phi^{-1}$.

\begin{restatable}[Two-component Guided Density]{corollary}{twocomponentdensity}
\label{cor:one-dimensional}
Choose $\mu_1,\mu_2\in\R$ satisfying $|\mu_1|=|\mu_2|$, and $\tau, \gamma >0$. Let $\Phi$ denote the standard normal CDF. For 
$p(x|c) = \frac12\Normal(x;\mu_1,\tau^2)+\frac12\Normal(x;\mu_2,\tau^2)$ and 
$p(x) = \Normal(x;0,\tau^2)$,  define,
\begin{align}
  F_{\gamma,t}(x)
  :=\frac12\sum_{k=1}^2
       \Phi\!\left(\frac{x-\gamma\alpha(t)\mu_k}{\tau\sqrt\gamma}\right), \quad 
  W_{\gamma,t}(x)&=\sqrt\gamma\exp\!\left(
       -\frac{\gamma-1}{2}[\Phi^{-1}(F_{\gamma,t}(x))]^2\right).
       \label{eq:explicit-weight}
\end{align}
Then the density $p_{\gamma,t}$ for $v_{\gamma,t}$ as defined in \Cref{eq:guided-density} is given by,
\begin{equation}
  p_{\gamma,t}(x|c)=\frac{W_{\gamma,t}(x)}{2}
       \sum_{k=1}^2\Normal(x;\gamma\alpha(t)\mu_k,\gamma\tau^2).
  \label{eq:explicit-density}
\end{equation}
\end{restatable}

\textbf{Interpretation:} \Cref{cor:one-dimensional} demonstrates that for $\gamma > 1$ the density $p_{\gamma,t}$ is no longer a mixture of Gaussians, but rather a mixture of \emph{skewed Gaussians} whose components have ``centered'' (using $u = x - \gamma \alpha(t) \mu_k$) densities of the form $g(u)\mathcal{N}(u; 0, \gamma \tau^2)$ and where the skewing function $g(u) = W_{\gamma,t}(u + \gamma \alpha(t)\mu_k)$ is a sigmoid-like function that tilts the components towards $y_{\gamma,t}(0)$. This suggests that while guidance causes magnification of the $\mu_{k} - \bar\mu$ component, it also simultaneously suppresses the outer tails of the distribution. We visualize $\hat{p}, p$, and $ W_{\gamma,t}$ for \Cref{cor:one-dimensional} in \Cref{fig:reweighting-1d}.

\textbf{Extension to Higher Dimensions}: We can generalize the above result in a straightforward manner to higher dimensions, provided that the mixture components additionally have symmetries around the direction of interest. We state this result informally, to give intuition about the behavior of $W_{\gamma,t}$.

\begin{theorem}[Guidance Under Symmetries, Informal, see \cref{thm:multidimensional-inward-reweighting}]
\label{thm:multidimensional-informal}
For simplicity, assume $\bar\mu = 0$ and $\sum_k\pi_k\mu_k=0$. For directions $e$ whose ray is fixed by symmetries of the weighted component means (which we formalize in \Cref{def:symmetrically-preserved-ray}), the flow $v_{\gamma, t}(re)$ lies along $e$ for all $r$ and $W_{\gamma,t}$ satisfies,
\begin{align*}
W_{\gamma, t}(r e) = \gamma^{n/2} \exp \left(-\frac{\gamma - 1}{2} [\Phi^{-1}(F_{\gamma,t}(r))]^2\right).
\end{align*}
for the CDF $F_{\gamma,t}$ of a continuous scalar random variable analogous to $\hat{p}_{\gamma,t}$ in \Cref{cor:one-dimensional} arising from the projected flow $\langle v_{\gamma,t}(re), e\rangle$.
\end{theorem}

\textbf{Interpretation}: As $\gamma \to \infty$, the magnification effect separates the components of $\hat{p}_{\gamma,t}$ and analogously flattens the 1-dimensional CDF $F_{\gamma,t}$. Therefore along these symmetric directions we can interpret $W_{\gamma,t}$ as behaving analogously to a step function in the large-$\gamma$ limit, with $\mu_k$- and $\Sigma$-dependent transition boundaries.

\subsection{Lower and upper bounds on the reweighting function in higher dimensions}
\label{sec:multidimensional-reweighting}

The above results suggest that although $W_{\gamma,t}(x)$ skews components towards $y_{\gamma,t}(0)$, this effect may not cause unbounded collapse to the ``center'' point, $y_{\gamma,t}(0)$. This intuition is also reflected in the original definition of $W_{\gamma,t}(x)$ in \Cref{eq:guided-density}, as $W_{\gamma,t}(x) \propto \exp(-\frac{\gamma - 1}{2\gamma}\|y^{-1}_{\gamma,t}(x)\|_{\Sigma^{-1}}^2)$, for which the coefficient in the exponent has the limit $(\gamma - 1)/(2\gamma) \to 1/2$ as $\gamma \to \infty$. 

However, this naive analysis does not account for potentially $\gamma$-dependent behavior in $y^{-1}_{\gamma,t}$ itself. In this section we derive bounds for $\|y^{-1}_{\gamma, t}(x)\|_{\Sigma^{-1}}$ that do not require solving for $y^{-1}$ globally and do not impose any additional restrictions on the problem setting. In particular, for the following theorem, we use local sensitivity analysis of $y^{-1}$ to both upper and lower bound $W_{\gamma,t}(x)$ with respect to $\|x - y_{\gamma,t}(0)\|_{\Sigma^{-1}}$, instead of $\|y^{-1}_{\gamma,t}(x)\|_{\Sigma^{-1}}$.

\begin{restatable}[Central Enhancement and Tail Suppression]{theorem}{centralouterreweighting}
\label{thm:reweighting-bounds}
Assume the setting of \Cref{thm:guided-density} with $n \geq 1$ and $\gamma>1$. Then, for every $t\in[0,1]$ and any $x \in \R^n$, the density factor $W_{\gamma,t}(x)$ satisfies,
\begin{align*}
    \exp \left(\hspace{-0.2em}-\frac{\gamma-1}{2\gamma}\|x - y_{\gamma,t}(0)\|_{\Sigma^{ \scaleto{-1}{3pt}}}^2\right)
    \leq \gamma^{-n/2} W_{\gamma,t}(x)
    \leq\exp \left(-\frac{\gamma-1}{2\gamma}e^{-\gamma R^2\alpha(t)^2}\|x - y_{\gamma,t}(0)\|_{\Sigma^{\scaleto{-1}{3pt}}}^2\right).
\end{align*}
Note that at $x = y_{\gamma,t}(0)$ these bounds match and recover that $W_{\gamma,t}(y_{\gamma,t}(0)) = \gamma^{n/2}$.
\end{restatable}

\textbf{Interpretation:} \Cref{thm:reweighting-bounds} suggests that $p_{\gamma,t}$ behaves like the product of the scaled mixture density $\hat{p}_{\gamma,t}$ with a density $q(x) \propto W_{\gamma,t}(x)$ with Gaussian density bounds with respect to $\|\cdot\|_{\Sigma^{-1}}$ for which the ``variance'' $\hat{\sigma}^2$ is in the range,
\begin{align*}
    1 \leq \frac{\gamma}{\gamma - 1} \leq \hat{\sigma}^2 \leq \frac{\gamma}{\gamma - 1} e^{\gamma R^2},
\end{align*}
for $\gamma > 1$ (note that $q(x)$ is not necessarily zero-mean and so is not sub-Gaussian in the technical sense). The lower bound $\hat \sigma \geq 1$ demonstrates that while $W_{\gamma,t}$ concentrates $p_{\gamma,t}$ around $y_{\gamma,t}(0)$, \textbf{the degree of distribution collapse has a $\gamma$-independent lower bound}. This intuition also matches our exact results in \Cref{cor:one-dimensional} and \Cref{thm:multidimensional-informal}, as the derived $W_{\gamma,t}$ do not collapse as $\gamma \to \infty$. 

Note that the bound $\hat{\sigma} \geq 1$ is expressed in terms of the shared $\|\cdot\|_{\Sigma^{-1}}$ metric. For the standard Euclidean metric $\|\cdot\|$ any equivalent bound would be data-dependent and significantly more complex when the covariance $\Sigma$ is not identical for all mixture components.
\section{Experiments}
\label{sec:experiments}

\begin{figure}[!t]
  \centering
  \includegraphics[width=\linewidth]{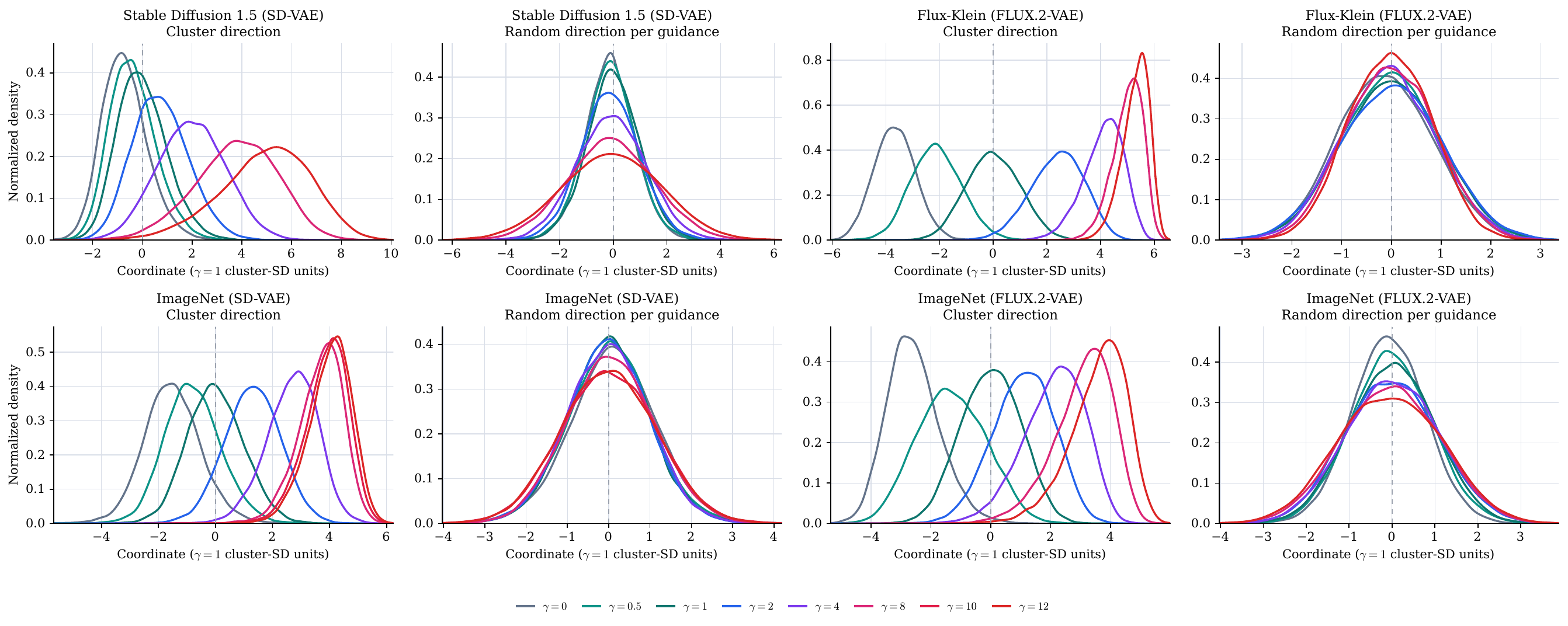}
  \vspace{-1.5em}
  \caption{Cluster direction and random-direction control densities across guidance scales. Each panel pairs a projection along the direction from the unconditional mean at $\gamma=0$ to the cluster mean at $\gamma=12$ with projection onto a random orthogonal direction (sampled independently per $\gamma$ value). At each guidance value, within-cluster standardized shapes are mapped to a shared mean and variance and aggregated across both conditioning values and clusters. Coordinates are expressed in $\gamma=1$ cluster standard-deviation units.}
  \label{fig:response-control-densities}
\end{figure}

\begin{figure}[!t]
  \centering
  \includegraphics[width=\linewidth]{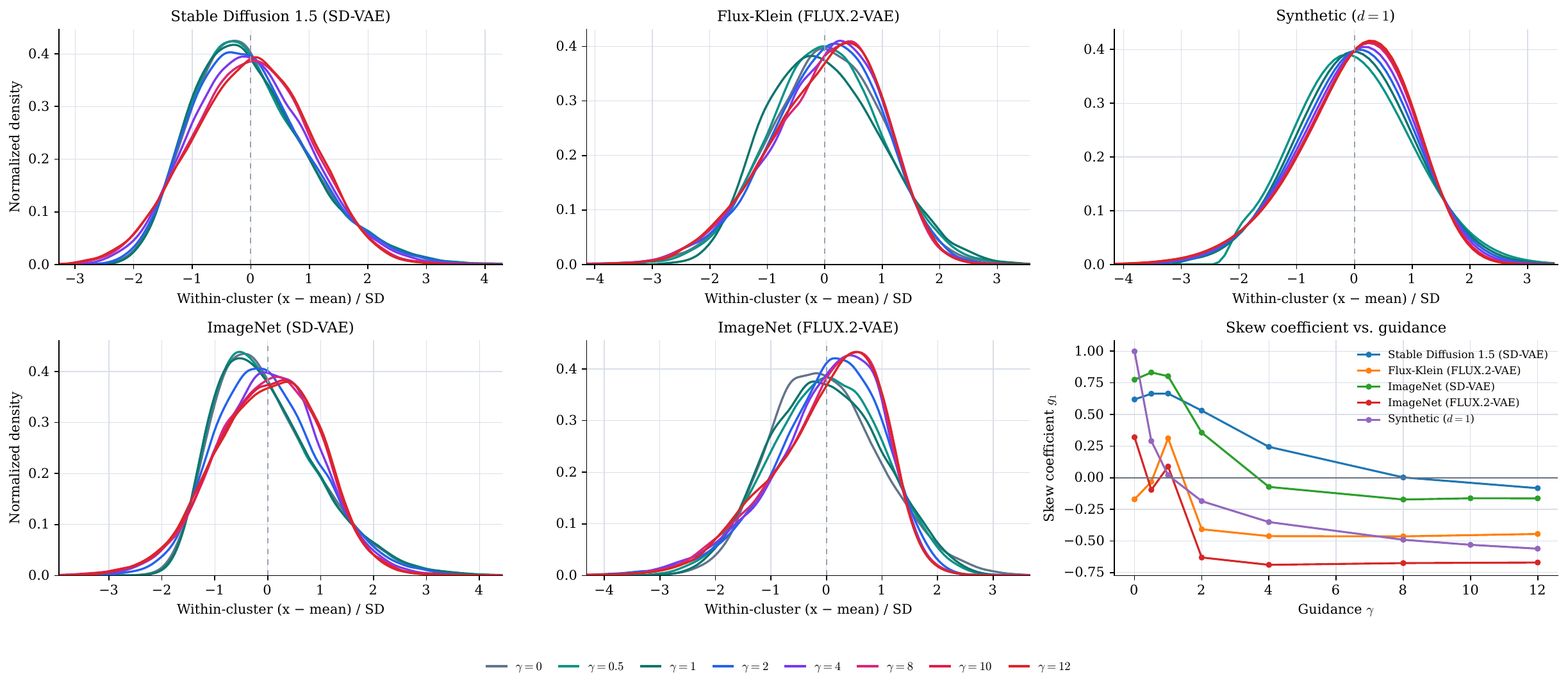}
  \vspace{-1em}
  \caption{Within-cluster standardized densities across guidance, using a per-$\gamma$ radial direction from each conditioning value's mean to each frozen cluster's mean. At each guidance value, each cluster's projections are normalized to zero mean and unit standard deviation before aggregation. Large $\gamma$ induces left-skewed distributions along this axis, matching the predicted behavior of the reweighting function $W_{\gamma,t}(x)$. The synthetic panel uses the same setup as \Cref{fig:reweighting-1d}.}
  \label{fig:standardized-response-densities}
\end{figure}

The existence of \emph{feature magnification} and \emph{mode-collapse} effects due to guidance have already been widely documented in prior literature \citep{jin2026stage, li2026towards}. In these experiments, we demonstrate that their effects are visible precisely along our predicted \emph{component-specific} axes and confirm that the degree of mode collapse is bounded, as suggested by our theory.

We conduct an empirical investigation into the behavior of latent image diffusion models in both text- and class-conditioned as well as different latent spaces. We use Stable Diffusion 1.5 and FLUX-Klein-9B for text-conditioned experiments and DiT models we trained with velocity prediction for class-conditional experiments on ImageNet. We use either noiseless DDIM or Euler-integration configurations to match the flow-based sampling ODE we analyze. See \Cref{app:experiment-details} for detailed experiment setup information. 

For our experiments, we approximate $\bar\mu$ by the average over $\gamma = 0$ samples (i.e. $\bar\mu \approx \mathbb{E}_{p_{\gamma=0}(x)}[x]$). For $y_{\gamma, 0}(0)$, we find that, with certain prompts and models, evaluating $y_{\gamma, 0}(0)$ directly produces highly artifacted images, as $x_1 = 0$ is out of distribution. As a more stable alternative we instead start the sampler from a \emph{low-frequency-zeroed} $x_1$: a draw $x_1 \sim \mathcal{N}(0,I)$ with the lowest $75\%$ of its 2D DCT frequencies along each spatial axis set to zero in every latent channel. This removes the coarse structure of $x_1$, which determines the global layout of the sample, while keeping the high-frequency statistics of the latent in-distribution for the model. This avoids most of the artifacts of $x_1=0$ while retaining the essential structure (see \Cref{fig:lowpass-ablation}).

\begin{figure}[!t]
  \centering
  \includegraphics[width=0.9\linewidth]{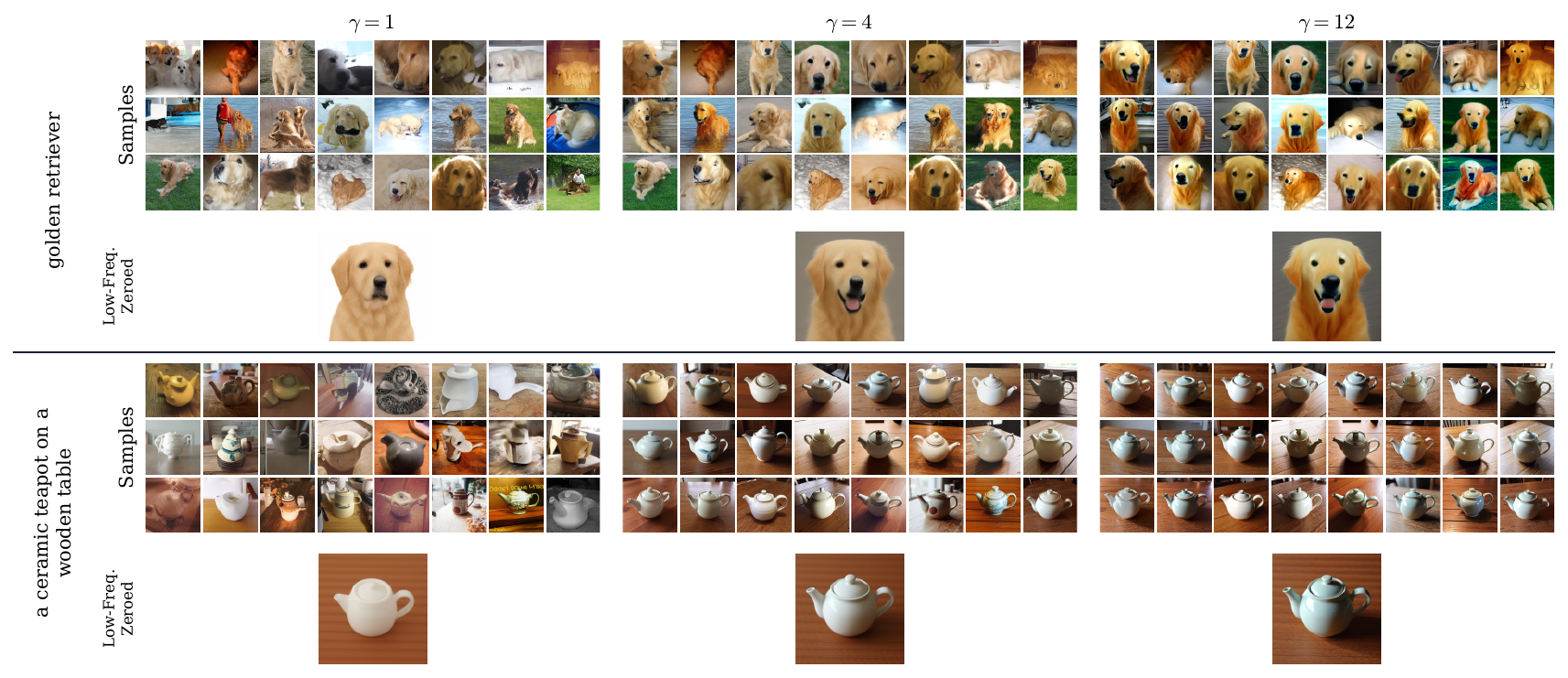}
  \caption{Outputs across guidance scales $\gamma=1,4,12$ for a selected ImageNet text prompt using FLUX latents (top) and FLUX-Klein-9B (bottom). The bottom row shows the low-frequency-zeroed $y_{\gamma,0}(0)$ while the top row shows samples matched across guidance scales by the initial value for $x_1$, demonstrating that increasing guidance concentrates samples around images with features similar to $y_{\gamma,0}(0)$, while some variation persists.}
  \vspace{-1em}
  \label{fig:dog-gallery}
\end{figure}

\begin{figure}[!t]
  \centering
  \includegraphics[width=\linewidth]{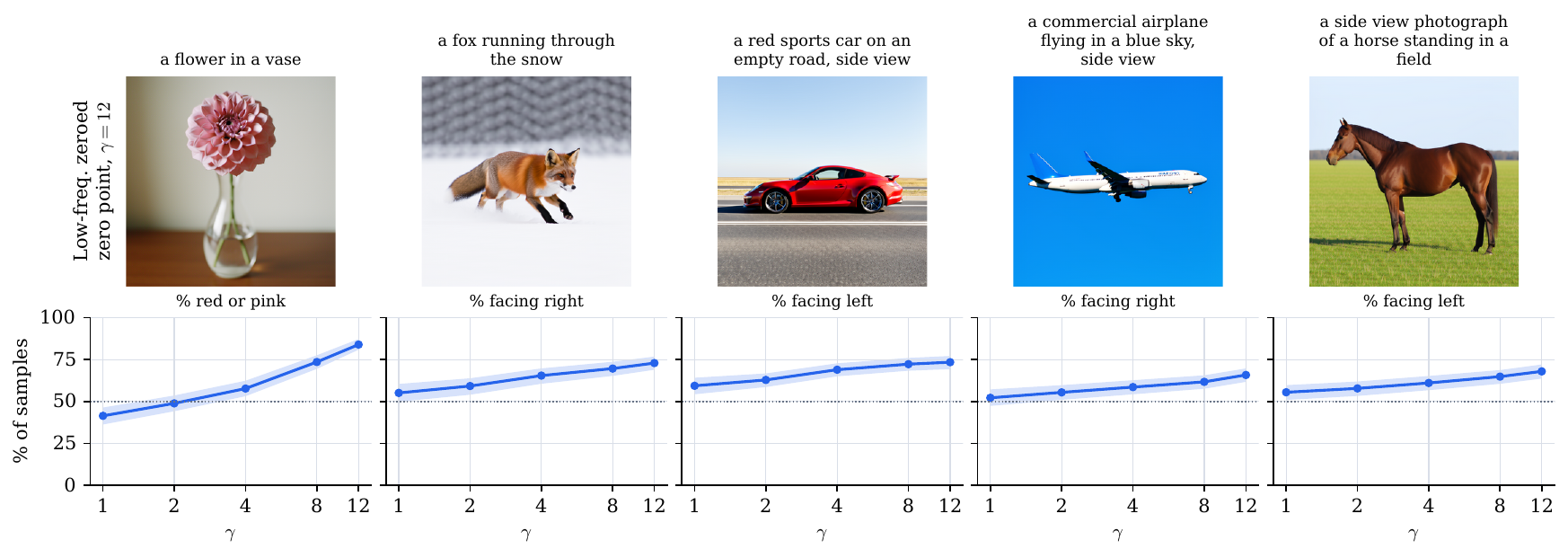}
  \caption{Mode collapse toward $y_{\gamma,0}(0)$ for FLUX-Klein-9B. Top: the low-frequency-zeroed $y_{\gamma,0}(0)$ at $\gamma=12$. Bottom: the percentage of samples in the mode that $y_{\gamma,0}(0)$ exhibits, as a function of $\gamma$, for prompts whose samples are near-evenly split between two modes at $\gamma=1$. Each point uses 512 initial noises shared across $\gamma$; shaded bands are 95\% bootstrap intervals. Modes are scored by a vision-language model (\Cref{app:zero-noise-collapse}).}
  \label{fig:zero-noise-collapse}
\end{figure}

\begin{figure}[!t]
  \centering
  \includegraphics[width=\linewidth]{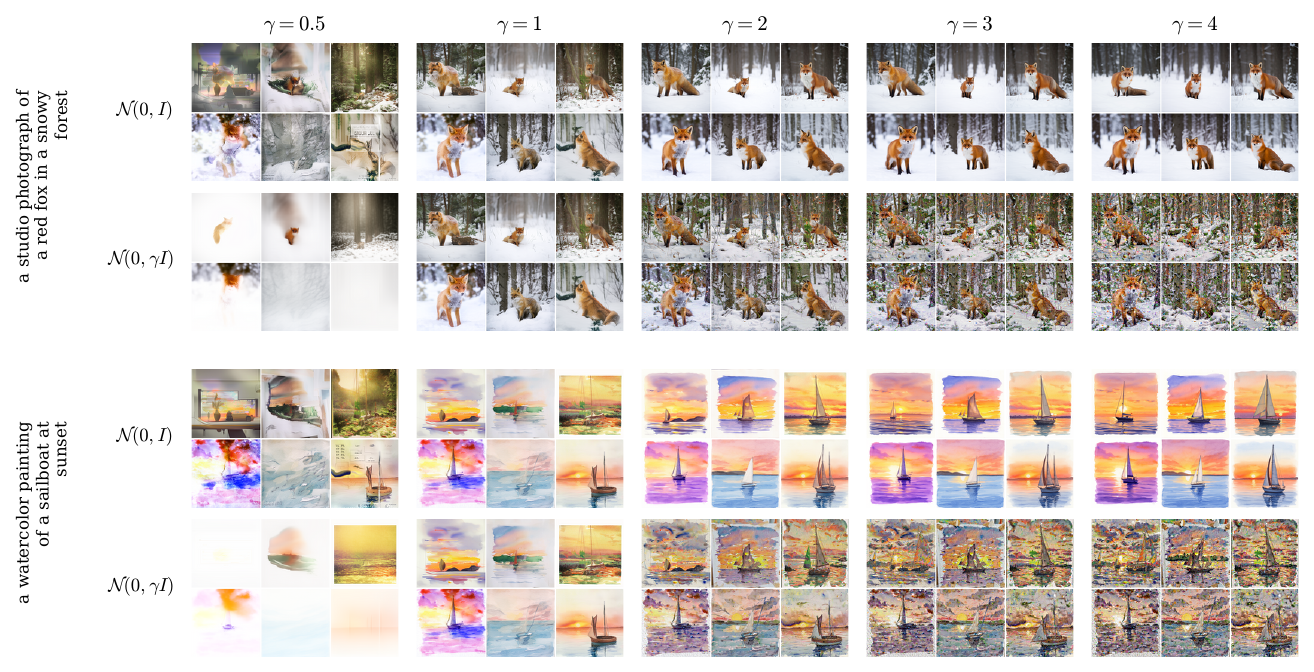}
  \caption{Samples at guidance scales $\gamma\in\{0.5,1,2,3,4\}$ started from the usual $x_1\sim\Normal(0,I)$ (top row of each prompt) and from $x_1\sim\Normal(0,\gamma I)$ (bottom row), for two FLUX Klein text prompts. Each cell shows six initial noises $\epsilon$ and the bottom rows use $\sqrt{\gamma}\,\epsilon$ for the same $\epsilon$, so the two rows coincide at $\gamma=1$. All four ImageNet classes and text prompts are shown in \Cref{fig:noise-scale-imagenet,fig:noise-scale-appendix}.}
  \label{fig:noise-scale-main}
\end{figure}

Note that for our analysis to hold we do \emph{not} require that $v_{\gamma,t}(y_{\gamma,t}(0))$ be well approximated by the learned model. The guided sampling procedure uses $x_1 \sim \mathcal{N}(0,I)$, which are in-distribution for the learned model, unlike $x_1 = 0$, for which one cannot necessarily expect a well-formed sample in any case. Our model predicts collapse to the true $y_{\gamma,t}(0)$ given the $p(x)$ and $p(x|c)$ distributions, \emph{even if} the learned $v_{\gamma,t}$ poorly captures that specific flow.

We compute these metrics in the latent space and use $K$-means clustering, with $K \in \{1, \ldots, 16\}$ chosen via Bayesian Information Criterion, to compute $\{\mu_k\}_{k=1}^{K}$ for $p(x|c)$. This allows us to empirically investigate our hypothesized guidance effects:
\begin{enumerate}[nosep,leftmargin=1.5em]
    \item In \Cref{fig:response-control-densities} we project each cluster along the direction from the estimated unconditional mean at $\gamma=0$ to the cluster mean at $\gamma=12$. We show that higher guidance values push samples along this axis, while randomly chosen axes see no net effect, precisely what our model predicts.
    \item In \Cref{fig:standardized-response-densities} we project each cluster along the direction from the condition mean to the cluster mean at each guidance scale and standardize the projections to zero mean and unit variance. Higher guidance induces left-skewed densities along this axis, consistent with the inward reweighting predicted by $W_{\gamma,t}(x)$.
    \item In \Cref{fig:dog-gallery} we visualize distributional collapse to a ``prototypical'' $y_{\gamma,0}(0)$ sample. We can clearly observe both concentration around $y_{\gamma,0}(0)$ in addition to persistence of other modes. For additional classes and samples, see \Cref{fig:imagenet-sample-gallery} and \Cref{fig:text-conditioned-sample-gallery}.
    \item In \Cref{fig:zero-noise-collapse} we test the predicted \emph{direction} of mode collapse. For prompts whose samples split near-evenly between two modes at $\gamma=1$ (e.g.\ the subject facing left or right, or the color of a flower), the share of samples in the mode exhibited by $y_{\gamma,0}(0)$ grows with $\gamma$. As predicted by our model, the degree of collapse is also bounded: while the share of $y_{\gamma,0}(0)$ samples increases, even at high $\gamma$, we still retain some diversity.
    \item In \Cref{fig:noise-scale-main} we show samples using $x_1 \sim \mathcal{N}(0, \gamma I)$ for $\gamma \in \{0.5, \ldots, 4\}$ as an analogue of $\hat{p}_{\gamma,0}$, where $\hat{p}_{\gamma,1}(x) = \mathcal{N}(0, I)$. While the samples exhibit significant artifacts as a result of the out-of-distribution initial state, we can observe that the matched $\mathcal{N}(0,I)$ versions have been ``normalized'' towards a more uniform layout and style.
\end{enumerate}

Together, these results support our theoretical account of guidance as combining directional magnification with inward reweighting, producing concentration without eliminating all distinct modes. We report additional experiments in \Cref{app:experiment-details}.

\section{Discussion and Conclusion}
\label{sec:conclusion}

Our results suggest that guidance can be understood as amplifying cluster-specific features and amplifying a specific ``central'' point. As our results are predicated on the assumption that the unconditional flow is a poor representation of the true unconditional distribution, this opens the possibility that advances in unconditional generation may conversely reduce the benefits of classifier-free guidance. For limitations of our model and extended discussion, see \Cref{sec:limitations}.

We believe our results should be viewed as complementary to existing results for linear flows, where classifier-free guidance yields a Gaussian with an altered mean and covariance. More broadly, an understanding of flow composition beyond the standard guidance scheme may allow the development of sampling methods with more fine-grained control of the sampling distribution.

\bibliographystyle{plainnat}
\bibliography{refs}

\clearpage
\appendix
%!TEX root = ../main.tex
\section{Extended Prior Work}
\label{app:prior_work}
\paragraph{Guidance and the tilted-distribution interpretation.}
In its original formulation, diffusion \citep{sohl2015deep, ho2020denoising, song2020score} was proposed as a mechanism for generating unconditional samples by reversing a stochastic random walk to iteratively ``denoise'' random data. The development of deterministic samplers such as DDIM \citep{song2020denoising} and more general theoretical frameworks such as flow-map matching and stochastic interpolants \citep{albergo2022building, albergo2025stochastic} has given diffusion a solid theoretical and empirical underpinning in this setting. However, prior to the introduction of \emph{classifier guidance} by \citet{dhariwal2021diffusion}, naive extension of this framework to the conditional setting underperformed contemporaneous GAN-based alternatives such as \citet{brock2018large}. Classifier guidance combines the scaled gradient of a noise-aware classifier with the output of a diffusion model with the aim of sampling from a density proportional to $p(x)\,p(c\mid x)^{\gamma}$. The subsequent development of \emph{classifier-free guidance} \citep{ho2022classifier} removed the dependency on an external classifier by extrapolating between a conditional and an unconditional diffusion model instead, implicitly targeting the \emph{tilted density} $p(x | c)^{\gamma}p(x)^{1 - \gamma}$. For simplicity, we use \emph{guidance} to refer primarily to the classifier-free variant of guidance throughout the rest of this work.

The fidelity--diversity
trade-off of guidance has been documented since inception: FID \citep{yu2021frechet}, which measures feature-distribution-level similarity, is optimal at small guidance weights while Inception Score \citep{salimans2016improved}, which more indirectly captures individual sample quality, keeps rising
\citep{dhariwal2021diffusion,ho2022classifier}, and large guidance weights $\gamma$ drive
pixel values to saturation~\citep{saharia2022photorealistic}. As a result of this phenomenon, there has been considerable interest in understanding the precise effect of guidance on the resulting sample distribution.

\paragraph{Guidance does not sample the tilted distribution.}
It was first observed by \citet{du2023reduce} that guidance does not faithfully sample from the tilted distribution. Although guidance targets the distribution $p(x|c)^{\gamma}p(x)^{1-\gamma}$ at each sampling step, the product of noised $p(x|c)$ and $p(x)$ distributions does not match the density of a noised $p(x|c)^{\gamma}p(x)^{1-\gamma}$ distribution, leading to intrinsic bias.
\citet{bradley2024classifier} show that guided DDPM and
guided DDIM produce different distributions, neither of which is the
$\gamma$-tilted density, and that in the SDE limit guidance acts as a predictor
for the conditional distribution alternating with a Langevin corrector for the
tilted one. \citet{chidambaram2024does} prove that, for mixtures of compactly
supported components, increasing the guidance
weight concentrates mass on the boundary of the conditional support rather than
on its modes. Recent analysis by \cite{jiang2026analytic} write
the distribution induced by the guided probability-flow ODE as a path-integral
correction to the marginal. Each of these expresses the guided density through
a correction accumulated along the sampling trajectory, rather than in closed
form.

\paragraph{Gaussian and Gaussian-mixture analyses.}
The closest antecedents to this work analyze guidance with exact scores
on Gaussian or Gaussian-mixture data. \citet{wu2024theoretical} prove formally that
guidance increases classification confidence and decreases the
differential entropy of the output for Gaussian mixtures.
\citet{li2026towards} decompose guidance with linear flows arising from a Gaussian model into a
mean shift towards the class-specific mean together with contrastive principal-component
terms that amplify class-specific directions and suppress directions shared
with the unconditional distribution while \citet{jin2026stage} analyze guidance effects across different time regimes. \citet{pavasovic2025classifier} characterize
the guided distribution for a two-cluster mixture in high dimension, finding an
overshot mean and shrunken variance at finite dimension that vanish as
$d\to\infty$.

These results provide monotonicity statements or stage-wise descriptions, but none gives an exact,
finite-dimensional density for the nonlinear flows that arise from multimodal
conditional distributions. Our restriction to mixtures with a common Mahalanobis radius is
what makes such a formula directly available.

\paragraph{Guidance beyond continuous diffusion.}
The same guidance-style extrapolation is now used well outside image diffusion, which
underlines the value of understanding its effect on the sampled distribution.
It improves prompt adherence in autoregressive language
models~\citep{sanchez2023stay}, transfers directly to flow
matching~\citep{zheng2023guided}, and is standard in masked and
autoregressive image generation~\citep{chang2023muse}, where it can also be
absorbed into training~\citep{chen2025visual,chen2025toward}. Novel one-step generative methods, such as MeanFlow \cite{geng2026mean}
and Sphere Encoders \citep{yue2026image} have devised guidance-style schemes for boosting empirical performance. Finally, guided samplers are routinely distilled
into single networks conditioned on the guidance
weight~\citep{meng2023distillation}, so that any systematic skew introduced by
guidance is inherited by the distilled model.

\section{Limitations and Extended Discussion}
\label{sec:limitations}

Our analysis relies on a particular model for both the unconditional and conditional distributions. It additionally requires the unconditional and conditional components to share the same covariance matrix, which is an unrealistic assumption, as we would expect the unconditional distribution to exhibit significantly higher covariance. Thus, we view our model as orthogonal to the intuition provided by existing result involving linear flows \citep{li2026towards, wu2024theoretical}, which can handle mismatched covariances between the unconditional and conditional distributions. With more precise analysis under a more general class of distributions, it may be possible to remove some of these structural limitations for our work.

Our results also presume that the learned diffusion model exactly matches the ground-truth flow for the modeled distributions. It does not capture effects such as generalization of the model itself to unobserved conditioning values, which may introduce its own biases \citep{pfrommer2026your, scarvelis2023closed}. Results in the generalization of diffusion models suggest that models may generalize to unseen conditioning values via a form of guidance analogous to classifier-free guidance, meaning that our results may provide insight into generalization behavior, although we treat such analysis as out-of-scope for this work.

Finally, it is unclear if our proposed model for guidance will apply given future advancements in the fidelity of samples generated by unconditional diffusion models or flows. It may be that for more expressive unconditional flows guidance does not confer fidelity benefits (as a more powerful unconditional model may invert the self-guidance effect observed in \cite{li2025self}). This may effectively serve as a litmus test as to whether our model captures the aspects of guidance that are relevant to the improved sample quality it currently produces.
%!TEX root = ../main.tex

\section{Deferred proofs}
\label{app:proofs}
\begin{restatable}[Finite Gaussian mixture]{proposition}{mixtureinterpolant}
\label{prop:mixture-interpolant}
Fix $\mu_1, \ldots, \mu_K \in \R^{d}$, covariances $\Sigma_1, \ldots, \Sigma_K \succ 0$, and some schedule $(\alpha, \sigma)$. Let $X_0$ be distributed according to,
\begin{equation}
  X_0\sim\sum_{k=1}^K\pi_k\Normal(\mu_k,\Sigma_k),
  \qquad \pi_k>0,
  \qquad \sum_{k=1}^K\pi_k=1,
\end{equation}
and $X_1 \sim \mathcal{N}(0,\Sigma)$. Define
$\Sigma_{k,t}:=\alpha(t)^2\Sigma_k+\sigma(t)^2\Sigma$. Then the density $p_t(x)$ of the interpolant $X_t$ and associated $v_t$ are given by
\begin{align}
  p_t(x)&=\sum_{k=1}^K\pi_k
  \Normal\bigl(x;\alpha(t)\mu_k,\Sigma_{k,t}\bigr), \\
  \label{eq:mixture-density}
    v_t(x) &= \sum_k w_k(t,x)
  \left[
  \dot{\alpha}(t)\mu_k+
  (\alpha(t)\dot{\alpha}(t)\Sigma_k+\sigma(t)\dot{\sigma}(t)\Sigma)
  \Sigma_{k,t}^{-1}(x-\alpha(t)\mu_k)
  \right].
\end{align}
where 
\[
w_k(t, x) := \frac{\frac{\pi_k}{\sqrt{\det(\Sigma_{k,t})}} e^{-\frac{1}{2}\|x - \alpha(t)\mu_k\|_{\Sigma_{k,t}^{-1}}^2}}{\sum_{j} \frac{\pi_j}{\sqrt{\det(\Sigma_{j,t})}} e^{-\frac{1}{2}\|x - \alpha(t)\mu_j\|_{\Sigma_{j,t}^{-1}}^2}}.
\]
\end{restatable}
\begin{proof}By \Cref{eq:velocity},
\begin{align*}
    v_t(x) &= \Exp[\dot{\alpha}(t)X_0 + \dot{\sigma}(t) X_1 | X_t = x] \\
    &= \Exp\left[\dot{\alpha}(t)X_0 + \frac{\dot{\sigma}(t)}{\sigma(t)}(X_t - \alpha(t)X_0) \middle| X_t = x\right] \\
    &= \frac{\dot{\sigma}(t)}{\sigma(t)}x + \left(\frac{\dot{\alpha}(t)\sigma(t) - \alpha(t)\dot{\sigma}(t)}{\sigma(t)}\right)\mathbb{E}[X_0 | X_t = x]
\end{align*}
Let $C$ be drawn from the categorical distribution over labels and $X_0$ be distributed such that $X_0 | C \overset{\mathrm{dist}}{=} \mathcal{N}(\mu_k,\Sigma_k)$. We can then write,
\begin{align*}
    \mathbb{E}[X_0 | X_t = x] &= \sum_{k=1}^{K} P(C =k | X_t = x)\mathbb{E}[X_0| X_t = x, C = k] \\
    &= \sum_{k=1}^{K} w_k(t, x) \int y \left(\frac{p(X_t = x| X_0 = y, C = k)p(X_0 = y | C=k)}{p(X_t = x | C = k)}\right) \rmd y \\
    &= \sum_{k=1}^{K} w_k(t, x) \int y C_k \exp\left(-\frac{1}{2\sigma(t)^2}\|x - \alpha(t)y\|_{\Sigma^{-1}}^2 - \frac{1}{2}\|y - \mu_k\|_{\Sigma_k^{-1}}^2 + \frac{1}{2}\|x - \alpha(t)\mu_k\|_{\Sigma_{k,t}^{-1}}^2\right) \rmd y \\
    &= \sum_{k=1}^{K} w_k(t, x) \int y C_k'(x) \exp\left(-\frac{\alpha(t)^2}{2\sigma(t)^2}\left\|\frac{1}{\alpha(t)}x - y\right\|_{\Sigma^{-1}}^2 - \frac{1}{2}\|y - \mu_k\|_{\Sigma_k^{-1}}^2\right)\rmd y \\
    &= \sum_{k=1}^{K} w_k(t, x) \int y C_k'(x) \exp\Big(-\frac{1}{2}(\alpha(t)^{-1}x - y)^\top\left(\frac{\sigma(t)^2}{\alpha(t)^2}\Sigma\right)^{-1}(\alpha(t)^{-1}x - y) \\
    &\hspace{15em} -\frac{1}{2}(y - \mu_k)^\top \Sigma_k^{-1} (y - \mu_k)\Big)\rmd y \\
\intertext{The product of two Gaussian densities $\mathcal{N}(x; \mu_A, \Sigma_A)$  and $\mathcal{N}(x; \mu_B, \Sigma_B)$ is proportional to the density$\mathcal{N}(x;  (\Sigma_A^{-1} + \Sigma_B^{-1})^{-1}(\Sigma_A^{-1}\mu_A + \Sigma_B^{-1}\mu_B), (\Sigma_A^{-1} + \Sigma_B^{-1})^{-1})$. In this case the normalization constant of the resulting density is precisely $C'_k(x)$, meaning we are integrating over a density with mean $(\sigma(t)^2\Sigma_{k}^{-1} + \alpha(t)^2\Sigma^{-1})^{-1}(\sigma(t)^2\Sigma_k^{-1} \mu_k + \alpha(t) \Sigma^{-1}x)$.}
    &= \sum_{k=1}^{K} w_k(t, x) \left(\Sigma_{k}^{-1} + \frac{\alpha(t)^2}{\sigma(t)^2}\Sigma^{-1}\right)^{-1}(\Sigma_k^{-1} \mu_k + \frac{\alpha(t)}{\sigma(t)^2}\Sigma^{-1} x) \\
    &= \sum_{k=1}^{K} w_k(t, x) \left(\sigma(t)^2\Sigma_{k}^{-1} + \alpha(t)^2\Sigma^{-1}\right)^{-1}(\sigma(t)^2\Sigma_k^{-1} \mu_k + \alpha(t)\Sigma^{-1} x)
\end{align*}
Substituting into the original expression, we have
\begin{align}
    v_t(x) = \sum_k w_k(t,x)
  \left[
  \dot{\alpha}(t)\mu_k+
  (\alpha(t)\dot{\alpha}(t)\Sigma_k+\sigma(t)\dot{\sigma}(t)\Sigma)
  \Sigma_{k,t}^{-1}(x-\alpha(t)\mu_k)
  \right]. \label{eq:mixture-velocity}
\end{align}
\end{proof}

\mahalanobismixtureflow*
\begin{proof}
This is just a direct specialization of \Cref{prop:mixture-interpolant}. Since $\Sigma_{k, t} = \Sigma$ and $\sigma^{2}(t) + \alpha^{2}(t) = 1$, $\sigma(t) \dot \sigma(t) = -\alpha(t) \dot \alpha(t)$, so the expression in \eqref{eq:mixture-velocity} simplifies to
\begin{align*}
  v_t(x) = \dot\alpha(t) \sum_{k} w_{k}(t, x) \mu_k,
\end{align*}
and the expression for the weights $w_k(t, x)$ follows by specializing to the case where $(\mu_k - \bar\mu)^\top \Sigma^{-1} (\mu_k - \bar\mu) = R^2$.
\end{proof}

\guideddensity*
\begin{proof} By \Cref{prop:mahalanobis-mixture-flow}, $v_{\gamma,t}(x)$ satisfies $\partial_t \hat{p}_{\gamma,t} + \nabla_x \cdot (v_{\gamma,t}\hat{p}_{\gamma,t}) = 0$ for
\begin{align*}
  \hat{p}_{\gamma,t}(x) := \sum_{k=1}^{K} \pi_k \mathcal{N}\bigl(x; \alpha(t)[\bar\mu+\gamma(\mu_k-\bar\mu)], \gamma \Sigma\bigr).
\end{align*}
Indeed, applying that proposition with means $\bar\mu+\gamma(\mu_k-\bar\mu)$, center $\bar\mu$, and covariance $\gamma\Sigma$ leaves the weights $w_k(t,x)$ unchanged and gives \eqref{eq:main_guided}. The auxiliary density has the boundary condition $\hat{p}_{\gamma,1}(x) = \Normal(x; 0, \gamma \Sigma)$.

Define $W_{\gamma,t}(x) := \frac{p_{\gamma,1}(y_{\gamma,t}^{-1}(x))}{\hat{p}_{\gamma,1}(y_{\gamma,t}^{-1}(x))}$ and $p_{\gamma,t}(x) := W_{\gamma,t}(x) \hat{p}_{\gamma,t}(x)$, such that $p_{\gamma,1}(x)=\Normal(x;0,\Sigma)$. Taking the ratio of the two Gaussian densities gives
\begin{align*}
  W_{\gamma,t}(x)=\gamma^{n/2}\exp\!\left(-\frac{\gamma-1}{2\gamma}\|y_{\gamma,t}^{-1}(x)\|_{\Sigma^{-1}}^2\right).
\end{align*}
Since $y_{\gamma,1}$ is the identity, the reweighted $p_{\gamma,t}$ satisfies the required boundary condition. To show that it still satisfies the continuity equation, write
\begin{align}
  \partial_t p_{\gamma,t}(x) &= \partial_t W_{\gamma,t}(x) \hat{p}_{\gamma,t}(x) + W_{\gamma,t}(x) \partial_t \hat{p}_{\gamma,t}(x) \\
  &=\partial_t W_{\gamma,t}(x) \hat{p}_{\gamma,t}(x) - W_{\gamma,t}(x) \nabla_x \cdot (\hat{p}_{\gamma,t} v_{\gamma,t}). \label{eq:cont-proof-inter}
\end{align}
Note that by cancellation of $y_{\gamma,t}$ and $y_{\gamma,t}^{-1}$, $W_{\gamma,t}(y_{\gamma,t}(x))$ is constant in $t$, so
\begin{align*}
  0 = \frac{\rmd}{\rmd t} W_{\gamma,t}(y_{\gamma,t}(x)) = \partial_t W_{\gamma,t}(y_{\gamma,t}(x)) + \nabla_x W_{\gamma,t}(y_{\gamma,t}(x)) \cdot v_{\gamma,t}(y_{\gamma,t}(x)).
\end{align*}
Since $y_{\gamma,t}$ is onto, $\partial_t W_{\gamma,t}(x) = - \nabla_x W_{\gamma,t}(x) \cdot v_{\gamma,t}(x)$ holds for all $x \in \R^n, t \in [0, 1]$. Substituting this into \eqref{eq:cont-proof-inter} and recognizing the product rule expression yields,
\begin{align*}
  \partial_t p_{\gamma,t}
  &=-\hat p_{\gamma,t}\,v_{\gamma,t}\cdot\nabla_x W_{\gamma,t}
    -W_{\gamma,t}\nabla_x\cdot(\hat p_{\gamma,t}v_{\gamma,t})
   =-\nabla_x\cdot(p_{\gamma,t}v_{\gamma,t}).
\end{align*}
Lastly, since normalization is preserved by the transport equation, since $p_{\gamma,1}(x) = \Normal(x;0,\Sigma)$, $p_{\gamma,t}(x)$ is a normalized density for all $t \in [0, 1]$.
\end{proof}

\twocomponentdensity*
\begin{proof}Note that $\hat{p}_{\gamma, t}(x) = \frac{1}{2} \mathcal{N}(x;\gamma\alpha(t)\mu_1, \gamma \tau^2) + \frac{1}{2}\mathcal{N}(x;\gamma\alpha(t)\mu_2, \gamma \tau^2)$ satisfies the continuity equation for $v_{\gamma, t}(x)$. Integrating the continuity equation yields
\begin{align*}
    0 &= \int_{-\infty}^x \left[\partial_t \hat{p}_{\gamma, t}(u) + \partial_{u}(\hat{p}_{\gamma, t}(u)v_{\gamma, t}(u))\right]\rmd u \\
    &= \partial_t F_{\gamma, t}(x) + \hat{p}_{\gamma, t}(x) v_{\gamma, t}(x) \\
    &= \partial_t F_{\gamma, t}(x) + [\partial_x F_{\gamma, t}(x)] v_{\gamma, t}(x).
\end{align*}
Thus for any given $x$, $\frac{\rmd}{\rmd t} F_{\gamma, t}(y_{\gamma, t}(x)) = \partial_t F_{\gamma, t}(y_{\gamma, t}(x)) + v_{\gamma, t}(y_{\gamma, t}(x)) \partial_x F_{\gamma, t}(y_{\gamma, t}(x)) = 0$ for all $t \in [0, 1]$, meaning that $F_{\gamma, t}(y_{\gamma, t}(x))$ is constant for all $t \in [0, 1]$. This yields the identity:
\begin{align*}
 \Phi\!\left(\frac{x}{\tau\sqrt{\gamma}}\right) = F_{\gamma,1}(y_{\gamma, 1}(x)) = F_{\gamma, t}(y_{\gamma, t}(x)).
\end{align*}
Thus,
\begin{align*}
 y_{\gamma,t}(x) &= F_{\gamma, t}^{-1}\!\left(\Phi\!\left(\frac{x}{\tau\sqrt{\gamma}}\right)\right), \\
 y_{\gamma,t}^{-1}(x) &= \tau\sqrt{\gamma} \Phi^{-1}(F_{\gamma, t}(x)).
\end{align*}
Since $\Sigma=\tau^2$, we have $\|y_{\gamma,t}^{-1}(x)\|_{\Sigma^{-1}}^2=\gamma[\Phi^{-1}(F_{\gamma,t}(x))]^2$. Substituting this into \Cref{eq:guided-density}, we get the result.
\end{proof}

\begin{definition}[Symmetrically preserved ray]
\label{def:symmetrically-preserved-ray}
Let $\nu_k:=\Sigma^{-1/2}\mu_k$ be the component means in whitened coordinates,
with weights $\pi_k$ and $\Sigma\succ0$. For a unit vector $\hat e$, write
\[
  e:=\frac{\Sigma^{1/2}\hat e}{\|\Sigma^{1/2}\hat e\|}.
\]
We call $\{re:r\geq0\}$ a \emph{symmetrically preserved ray} if there is a
group $H$ of orthogonal transformations preserving the weighted distribution
of the $\nu_k$, whose common fixed subspace is $\operatorname{span}\{\hat e\}$.
Explicitly, require
\begin{align}
  \sum_k\pi_k\delta_{Q\nu_k}
    &=\sum_k\pi_k\delta_{\nu_k}\qquad(Q\in H), \label{eq:ray-symmetry}\\
  \operatorname{span}\{\hat e\}
    &=\{z\in\R^n:Qz=z\text{ for every }Q\in H\}.
\end{align}
\end{definition}

\begin{restatable}[Analysis along symmetrically preserved rays]{theorem}{inwardtailpreference}
\label{thm:multidimensional-inward-reweighting}
Assume the setting of \Cref{thm:guided-density}, with $\bar\mu=0$, $R>0$,
$\gamma>1$, $\sum_k\pi_k\mu_k=0$ and $\Sigma\succ0$. Let $\{re:r\geq0\}$
be a symmetrically preserved ray as in \Cref{def:symmetrically-preserved-ray}, write $\lambda:=e^\top\Sigma^{-1}e$ and define the scalar velocity $\hat v_{\gamma, t}(r) := \langle e, v_{\gamma,t}(re)\rangle$. Then,
\begin{enumerate}
    \item The ray $\{ r e: r \geq 0\}$ is invariant under the guided flow $v_{\gamma,t}$ for all $\gamma \geq 0, t \in [0,1]$.
    \item Let $q_{\gamma,t}$ be the unique probability density on $\R$ satisfying,
    \begin{equation}
      \partial_t q_{\gamma,t}
      +\partial_r(\hat v_{\gamma,t}q_{\gamma,t})=0,
      \qquad q_{\gamma,1}(r)=\Normal(r;0,\gamma/\lambda).
      \label{eq:ray-scalar-transport}
    \end{equation}
    Define $F_{\gamma, t}(r) := \int_{-\infty}^{r} q_{\gamma,t}(u) \rmd u$. Then for every $t \in [0, 1]$ and $r \in \R$ the weighting function $W_{\gamma, t}(x)$ for $p_{\gamma,t}(x|c)$ as defined in \eqref{eq:explicit-weight} satisfies,
    \begin{equation}
      W_{\gamma,t}(re)=\gamma^{n/2}\exp\!\left(
        -\frac{\gamma-1}{2}
        \bigl[\Phi^{-1}(F_{\gamma,t}(r))\bigr]^2\right).
      \label{eq:ray-weight-monotonicity}
    \end{equation}
    Moreover $F_{\gamma,t,e}(0)=1/2$ and $W_{\gamma,t}(re)$ is strictly decreasing in $r>0$.
    \item Provided the orthogonal components of the whitened means have equal magnitudes, $q_{\gamma,t}$ can be written as a Gaussian mixture. Explicitly, let $a_k:=\hat e^\top\nu_k$ and suppose that for some $B \geq 0$ and all $k$, $\|\nu_k-a_k\hat e\|^2=B^2$. Then,
\begin{equation}
  q_{\gamma,t}(r)=\sum_k\pi_k
  \Normal\!\left(r;\frac{\gamma\alpha(t)a_k}{\sqrt\lambda},
    \frac{\gamma}{\lambda}\right).
  \label{eq:ray-mixture-density}
\end{equation}
\end{enumerate}
\end{restatable}
\begin{proof}
\textbf{Proof of 1.}
In whitened coordinates $z=\Sigma^{-1/2}x$, the guided velocity is
\begin{equation}
  \widetilde v_{\gamma,t}(z)
  =\gamma\dot\alpha(t)
    \frac{\sum_k\pi_k e^{\alpha(t)z^\top\nu_k}\nu_k}
         {\sum_k\pi_k e^{\alpha(t)z^\top\nu_k}}.
  \label{eq:whitened-guided-velocity}
\end{equation}
Let $H$ be a symmetry group as in \Cref{def:symmetrically-preserved-ray}.
The invariance in \eqref{eq:ray-symmetry} implies
$\widetilde v_{\gamma,t}(Qz)=Q\widetilde v_{\gamma,t}(z)$ for every
$Q\in H$. Along $z=r \hat e$, applying both properties implies that
\begin{align*}
 \widetilde v_{\gamma, t}(r \hat e) = \widetilde v_{\gamma, t}(Q r \hat e) = Q \widetilde v_{\gamma, t}(r \hat e).
\end{align*}
Since this holds for every $Q\in H$, the common fixed-subspace assumption
implies that $\widetilde v_{\gamma,t}(r\hat e)$ lies in
$\operatorname{span}\{\hat e\}$. Converting back to the original
coordinates gives $v_{\gamma,t}(re)\in\operatorname{span}\{e\}$.

The restriction of the flow to this line is the flow of a scalar ODE.
With $\hat v_{\gamma,s}(r)=\langle e,v_{\gamma,s}(re)\rangle$, fix $t$
and define $\rho_{t,s}$ as the transport map arising from integrating the scalar velocity from $t$ to $s$ and $h_t(r):=\rho_{t,1}(r)$ as the transport from $t$ to $1$.
\begin{equation}
  \frac{\rmd}{\rmd s}\rho_{t,s}(r)=\hat v_{\gamma,s}(\rho_{t,s}(r)),
  \qquad \rho_{t,t}(r)=r,\qquad h_t(r):=\rho_{t,1}(r).
\end{equation}
The scalar velocity $\hat{v}_{\gamma,t}$ and its spatial derivative are bounded, so this ODE
has a unique smooth flow given an initial condition. Since $v_{\gamma,t}(r e) \in \mathrm{span}\{e\}$ we can write,
\begin{equation}
  y_{\gamma,t}^{-1}(re)=h_t(r)e,
  \qquad
  \label{eq:ray-characteristic}
\end{equation}

The mean-zero assumption implies $v_{\gamma,t}(0)=0$ for all $t$ and thus $h_t(0)=0$. Since $h_t$ is injective and $h_t'(0) > 0$ (as will be shown below), $h_t(r) > 0$ for $r > 0$, meaning that the positive ray $\{r e \}$ is preserved.

\textbf{Proof of 2. Having defined $h_t(r)$, we proceed analogously to \Cref{cor:one-dimensional}.} The density $q_{\gamma,t}$ in \eqref{eq:ray-scalar-transport} is the
transport of its Gaussian density by this scalar flow. Integrating the continuity equation
for $q_{\gamma,t}$ from $-\infty$ to $r$ (as in the proof of \Cref{cor:one-dimensional}), yields,
\begin{equation}
  \partial_t F_{\gamma,t}
  +\hat v_{\gamma,t}\,\partial_r F_{\gamma,t}=0.
\end{equation}
Observe that $\frac{\rmd}{\rmd s} F_{\gamma,s}(\rho_{t,s}(r)) = \partial_s F_{\gamma,s}(\rho_{t,s}(r)) + \hat v_{\gamma,s}\,\partial_r F_{\gamma,s}(\rho_{t,s}(r)) = 0$ which means that $F_{\gamma,s}(\rho_{t,s}(r))$ is constant for all $s$. Using $s = t$ (for which $\rho_{s,s}$ is identity) and $s = 1$ (for which $F_{\gamma,1}$ is the CDF of $\Normal(0, \gamma/\lambda)$), we get,
\begin{equation}
  F_{\gamma,t}(r)
  =F_{\gamma,1}(\rho_{t,1}(r))
  =\Phi\!\left(\sqrt{\lambda/\gamma}\,h_t(r)\right).
  \label{eq:ray-cdf-characteristic}
\end{equation}
In particular, since $h_t(0) = 0$, $F_{\gamma,t}(0)=1/2$. By
\eqref{eq:guided-density} and \eqref{eq:ray-characteristic},
\begin{equation}
  W_{\gamma,t}(re)
  =\gamma^{n/2}\exp\!\left(
    -\frac{\gamma-1}{2\gamma}\lambda h_t(r)^2\right).\label{eq:pre-sub}
\end{equation}
Inverting \eqref{eq:ray-cdf-characteristic} and substituting into \eqref{eq:pre-sub} proves \eqref{eq:ray-weight-monotonicity},
\begin{align*}
  W_{\gamma,t}(re)=\gamma^{n/2}\exp\!\left(-\frac{\gamma-1}{2}
        \bigl[\Phi^{-1}(F_{\gamma,t}(r))\bigr]^2\right).
\end{align*}
Since the density of $q_{\gamma,1}$ is strictly positive and $v_{\gamma,t}$ is continuous, $q_{\gamma,t}$ also has strictly positive density for all $t$ and the associated CDF, $F_{\gamma,t}$, is strictly increasing. As both $\Phi^{-1}$ and $F_{\gamma,t}$ are strictly increasing, $W_{\gamma,t}(r e)$ must be strictly increasing for $r < 0$ and decreasing for $r > 0$. Note that this also implies that $h_t'(0) > 0$, completing the requirement we deferred from the proof of the first condition.

\textbf{Proof of 3.} Finally, suppose that $B \geq 0$ and all $k$, $\|\nu_k - a_k \hat e\|^2 = B^2$ holds. Since
$\Sigma^{-1/2}e=\sqrt\lambda\,\hat e$, writing out the projected scalar velocity, the weights for the orthogonal components are the same for all $k$ and therefore we obtain,
\begin{equation}
  \hat v_{\gamma,t}(r)=\dot\alpha(t) 
  \frac{\sum_k\pi_k e^{\alpha(t)\sqrt\lambda\,r a_k}\left[\frac{\gamma a_k }{\sqrt{\lambda}}\right]}
       {\sum_k\pi_k e^{\alpha(t)\sqrt\lambda\,r a_k}}.
\end{equation}
We can directly observe from \Cref{prop:mahalanobis-mixture-flow} that this is the flow for the mixture:
\begin{align*}
    q_{\gamma,t}(r) = \sum_{k=1}^{K} \pi_k \Normal\left(r, \frac{\gamma \alpha(t) a_k}{\sqrt{\lambda}}, \frac{\gamma}{\lambda}\right).
\end{align*}
\end{proof}

\centralouterreweighting*
\begin{proof}
For simplicity we can additionally assume $\bar \mu = 0$, as we can simply consider the centered coordinate space $x - \alpha(t) \bar\mu$ in which the mixture components lie on an ellipse centered at $0$ with $\|\cdot\|_{\Sigma^{-1}}$ radius $R$ and where guidance is equivalent to scaling the conditional flow.

\textbf{High-level strategy}: For this proof, we will show that,
\begin{align*}
    e^{-\gamma R^2 \alpha(t)^2/2}\|x - y_{\gamma,t}(0)\|_{\Sigma^{-1}}\leq\|y_{\gamma,t}^{-1}(x)\|_{\Sigma^{-1}}\leq\|x - y_{\gamma,t}(0)\|_{\Sigma^{-1}},
\end{align*}
holds for all $x$ and $t \in [0, 1]$. To construct these bounds, we use that $y_{\gamma,t}^{-1}(y_{\gamma,t}(0))=0$ and then bound $y_{\gamma,t}^{-1}(x)$ via sensitivity analysis of its spatial derivative, combined with integration along the segment from $y_{\gamma,t}(0)$ to $x$ for the upper bound and from $0$ to $y_{\gamma,t}^{-1}(x)$ for the lower bound.

Let $D$ denote differentiation with respect to the spatial argument. Choose any $z$ and tangent $u$ and set $h_s:=D y_{\gamma,s}(z)u$. Write $\langle a,b\rangle_{\Sigma^{-1}}:=a^\top\Sigma^{-1}b$ and $m_s(z):=\sum_k w_k(s,z)\mu_k$.  Differentiating the flow equation with respect to $z$ gives $\dot h_s=D v_{\gamma,s}(y_{\gamma,s}(z))h_s$ and $h_1=u$. Consequently,
\begin{align*}
   \frac{\rmd}{\rmd s}\|h_s\|_{\Sigma^{-1}}^2 =2\langle h_s,D v_{\gamma,s}(y_{\gamma,s}(z))h_s\rangle_{\Sigma^{-1}}.
\end{align*}
Differentiating the weights in \eqref{eq:common-radius-velocity} in any direction $h$ gives
\begin{align*}
    D w_k(s,z)h
    &=\alpha(s)w_k(s,z)\langle\mu_k-m_s(z),h\rangle_{\Sigma^{-1}},\\
    \langle h,D v_{\gamma,s}(z)h\rangle_{\Sigma^{-1}}
    &=\gamma\dot\alpha(s)\alpha(s)\left[\sum_k w_k(s,z)\langle\mu_k,h\rangle_{\Sigma^{-1}}^2-\langle m_s(z),h\rangle_{\Sigma^{-1}}^2\right].
\end{align*}
Dropping the negative term and applying Cauchy--Schwarz yields
\begin{align*}
    \sum_k w_k(s,z)\langle\mu_k,h\rangle_{\Sigma^{-1}}^2
    \leq\sum_k w_k(s,z)\|\mu_k\|_{\Sigma^{-1}}^2\|h\|_{\Sigma^{-1}}^2=R^2\|h\|_{\Sigma^{-1}}^2.
\end{align*}
Since 
\[
\sum_k w_k(s,z)\langle\mu_k,h\rangle_{\Sigma^{-1}}^2-\langle m_s(z),h\rangle_{\Sigma^{-1}}^2 = \sum_k w_k(s, z)\langle\mu_k-m_s(z),h\rangle_{\Sigma^{-1}}^2 \geq 0,
\]
and $\dot\alpha(s)\leq0, \alpha(s)\geq0$, this implies,
\begin{align*}
    -\gamma|\dot\alpha(s)|\alpha(s)R^2\|h\|_{\Sigma^{-1}}^2
    \leq\langle h,D v_{\gamma,s}(z)h\rangle_{\Sigma^{-1}}\leq 0.
\end{align*}
Therefore,
\begin{align*}
    -2\gamma|\dot\alpha(s)|\alpha(s)R^2\|h_s\|_{\Sigma^{-1}}^2
    \leq \frac{\rmd}{\rmd s} \|h_s\|^2_{\Sigma^{-1}} \leq 0.
\end{align*}
Note that,
\begin{align*}
    \frac{\rmd}{\rmd s} \left[e^{2\int_t^s\gamma|\dot\alpha(r)|\alpha(r)R^2\rmd r}\|h_s\|_{\Sigma^{-1}}^2\right] =e^{2\int_t^s\gamma|\dot\alpha(r)|\alpha(r)R^2\rmd r} \left[ 2\gamma|\dot\alpha(s)|\alpha(s)R^2\|h_s\|_{\Sigma^{-1}}^2 +  \frac{\rmd}{\rmd s} \|h_s\|^2_{\Sigma^{-1}}\right].
\end{align*}
Thus the right inequality implies that $\|h_s\|_{\Sigma^{-1}}^2$ is non-increasing, while the left implies that $e^{2\int_t^s\gamma|\dot\alpha(r)|\alpha(r)R^2\rmd r}\|h_s\|_{\Sigma^{-1}}^2$ is non-decreasing. Since $\dot\alpha\leq0$ and $\alpha(1)=0$, let $L_t:=\int_t^1\gamma|\dot\alpha(r)|\alpha(r)R^2\rmd r = \gamma R^2 \alpha(t)^2/2$. Comparing their values at $t$ and $1$ yields,
\begin{align*}
    e^{-2L_t}\|h_t\|_{\Sigma^{-1}}^2\leq\|h_1\|_{\Sigma^{-1}}^2\leq\|h_t\|_{\Sigma^{-1}}^2,
\end{align*}
Taking the square root, rearranging, and using that $h_t = D y_{\gamma,t}(z)u$, $h_1 = u$ gives the inequality:
\begin{align}
    \|u\|_{\Sigma^{-1}}\leq\|D y_{\gamma,t}(z)u\|_{\Sigma^{-1}}\leq e^{L_t}\|u\|_{\Sigma^{-1}}. \label{eq:inter-inequality}
\end{align}
\textbf{Converting the bounds to the inverse mapping}. To complete our argument, ultimately we want bounds on $\|D y_{\gamma,t}^{-1}(z)u\|_{\Sigma^{-1}}$ instead of $\|D y_{\gamma,t}(x)u\|_{\Sigma^{-1}}$. Differentiating $y_{\gamma,t}(y_{\gamma,t}^{-1}(x))=x$ by the chain rule gives that for all $\tilde u$,
\begin{align*}
    D y_{\gamma,t}(y_{\gamma,t}^{-1}(x))D y_{\gamma,t}^{-1}(x)\tilde u= \tilde u.
\end{align*}
Applying the lower derivative bound of \eqref{eq:inter-inequality} at the point $z = y_{\gamma,t}^{-1}(x)$ and in the direction $u = D y_{\gamma,t}^{-1}(x)\tilde u$ gives
\begin{align*}
    \|D y_{\gamma,t}^{-1}(x)\tilde u\|_{\Sigma^{-1}}
    &\leq\|D y_{\gamma,t}(y_{\gamma,t}^{-1}(x))D y_{\gamma,t}^{-1}(x)\tilde u\|_{\Sigma^{-1}}
    =\|\tilde u\|_{\Sigma^{-1}}.
\end{align*}
For all $t$, we have $y_{\gamma,t}^{-1}(y_{\gamma,t}(0))=0$. Therefore we can bound via integration along the segment $\{\lambda x + (1 - \lambda)y_{\gamma,t}(0) : \lambda \in [0, 1]\}$ from $y_{\gamma,t}(0)$ to $x$.
\begin{align*}
    y_{\gamma,t}^{-1}(x)
    &= y^{-1}_{\gamma, t}(x) - y^{-1}_{\gamma, t}(y_{\gamma,t}(0)) \\
    &= \int_0^1 D y_{\gamma,t}^{-1}(\lambda x + (1 - \lambda)y_{\gamma,t}(0))[x - y_{\gamma,t}(0)]\,\rmd \lambda,\\
    \|y_{\gamma,t}^{-1}(x)\|_{\Sigma^{-1}}
    &\leq\int_0^1\|D y_{\gamma,t}^{-1}(\lambda x + (1 - \lambda)y_{\gamma,t}(0))[x - y_{\gamma,t}(0)]\|_{\Sigma^{-1}}\,\rmd \lambda \\
    &\leq\int_0^1\|x - y_{\gamma,t}(0)\|_{\Sigma^{-1}}\,\rmd \lambda \\
    =\|x -y_{\gamma,t}(0)\|_{\Sigma^{-1}}.
\end{align*}
For the other inequality, apply the same argument to $y_{\gamma,t}$ along the segment from $0$ to $y_{\gamma,t}^{-1}(x)$, using its upper derivative bound:
\begin{align*}
    x - y_{\gamma,t}(0) &=y_{\gamma,t}(y_{\gamma,t}^{-1}(x))-y_{\gamma,t}(y_{\gamma,t}^{-1}(y_{\gamma,t}(0))) \\
    &= y_{\gamma,t}(y_{\gamma,t}^{-1}(x))-y_{\gamma,t}(0) \\
    &=\int_0^1 D y_{\gamma,t}(\lambda y_{\gamma,t}^{-1}(x))y_{\gamma,t}^{-1}(x)\,\rmd \lambda,\\
    \|x - y_{\gamma,t}(0)\|_{\Sigma^{-1}}
    &\leq\int_0^1\|D y_{\gamma,t}(\lambda y_{\gamma,t}^{-1}(x))y_{\gamma,t}^{-1}(x)\|_{\Sigma^{-1}}\,\rmd \lambda\\
    &\leq\int_0^1 e^{L_t}\|y_{\gamma,t}^{-1}(x)\|_{\Sigma^{-1}}\,\rmd \lambda
    =e^{L_t}\|y_{\gamma,t}^{-1}(x)\|_{\Sigma^{-1}}.
\end{align*}
Dividing the last inequality by $e^{L_t}$ and combining the two bounds yields,
\begin{align*}
    e^{-\gamma R^2 \alpha(t)^2/2}\|x - y_{\gamma,t}(0)\|_{\Sigma^{-1}}\leq\|y_{\gamma,t}^{-1}(x)\|_{\Sigma^{-1}}\leq\|x - y_{\gamma,t}(0)\|_{\Sigma^{-1}}.
\end{align*}
Squaring these bounds and substituting into \eqref{eq:guided-density} therefore yields,
\begin{align*}
    \gamma^{n/2}\exp\!\left(-\frac{\gamma-1}{2\gamma}\|x - y_{\gamma,t}(0)\|_{\Sigma^{-1}}^2\right)
    \leq W_{\gamma,t}(x)
    \leq\gamma^{n/2}\exp\!\left(-\frac{\gamma-1}{2\gamma}e^{-\gamma R^2\alpha(t)^2}\|x-y_{\gamma,t}(0)\|_{\Sigma^{-1}}^2\right).
\end{align*}
\end{proof}
\clearpage
\section{Experiment Details}
\label{app:experiment-details}

\textbf{Models and Sampling.}
We use pretrained Stable Diffusion 1.5 and FLUX.2-klein-base-9B for text-conditioned generation. For the ImageNet experiments we trained class-conditional DiT models ourselves with velocity ($v$) prediction: a DiT-B/2 on SD-VAE latents and a DiT-B/1 on FLUX.2-VAE latents. We use their EMA weights for sampling.
We generate text-conditioned images at $512\times512$ and ImageNet images at $256\times256$.

Each image experiment uses 10,000 samples per condition and guidance scale, with the same initial noise across scales. The density comparisons use $\gamma\in\{0,0.5,1,1.5,2,4,8,12\}$, with additional values $6.5$ and $10$ for the ImageNet (SD-VAE) experiments. We use four ImageNet classes: golden retriever, Arctic fox, fire engine, and strawberry. The text prompts are ``a studio photograph of a red fox in a snowy forest,'' ``a watercolor painting of a sailboat at sunset,'' ``a futuristic city street at night,'' and ``a ceramic teapot on a wooden table.''

\textbf{Noise Schedules.} For all models we use their ``native'' noise schedules as used by upstream repositories and configuration.
The SD1.5 experiments use a variance-preserving schedule with $\alpha(t_j)=\sqrt{\prod_{i=1}^{j}(1-\beta_i)}$ and $\sigma(t_j)=\sqrt{1-\alpha(t_j)^2}$ on 1,000 training noise levels, with $\sqrt{\beta_i}$ linearly spaced between $\sqrt{0.00085}$ and $\sqrt{0.012}$, and 50-step deterministic DDIM sampling.
The FLUX Klein and both ImageNet experiments use linear flow matching, $\alpha(t)=1-t$ and $\sigma(t)=t$. FLUX Klein uses 20 Euler steps with the model's default exponential time shift. The ImageNet experiments use 50 Euler steps at uniformly spaced times from $1$ to $t_{\min}$, followed by $0$, with $t_{\min}=0.04$ for SD-VAE and $t_{\min}=0.18$ for FLUX.2 latents.

\textbf{Cluster Projections.}
We use half the samples for fitting and half for evaluation. For each condition, we reduce flattened $\gamma=1$ latents to 32 principal components and fit k-means, selecting 1--16 clusters by spherical-Gaussian BIC. Assignments remain fixed across guidance, and projection directions use the fitting half. We retain clusters with at least 24 samples in each half.
The cluster direction in \Cref{fig:response-control-densities} points from the unconditional mean at $\gamma=0$ to the cluster mean at $\gamma=12$. The control direction is sampled independently at each scale, orthogonal to this displacement. For \Cref{fig:standardized-response-densities}, we instead project from the condition mean toward the cluster mean at each scale.
We standardize the held-out projections within each cluster and pool them, giving equal weight to conditions and to clusters within each condition. We estimate densities with Gaussian KDE at bandwidth $0.15$, correcting for the variance added by smoothing. The cluster-direction plots restore the weighted cluster mean and within-cluster variance, expressed relative to $\gamma=1$. The radial plots retain zero mean and unit variance, with skewness computed as the pooled third moment.

\textbf{Low-Frequency-Zeroed $y_{\gamma,0}(0)$.}
We compute the 2D DCT-II of each channel of $x_1 \sim \mathcal{N}(0,I)$ over the spatial latent grid, set the lowest $\mathrm{round}(f H) \times \mathrm{round}(f W)$ coefficients to zero, invert the transform, and run the guided sampler from the result without renormalizing. We use $f=0.75$ throughout, i.e.\ the lowest $24\times24$ frequencies of the $32\times32$ FLUX-Klein latent grid; $f=1$ recovers $x_1=0$. \Cref{fig:lowpass-ablation} shows the effect of increasing $f$ for a single draw of $x_1$.

\begin{figure}[!ht]
  \centering
  \includegraphics[width=\linewidth]{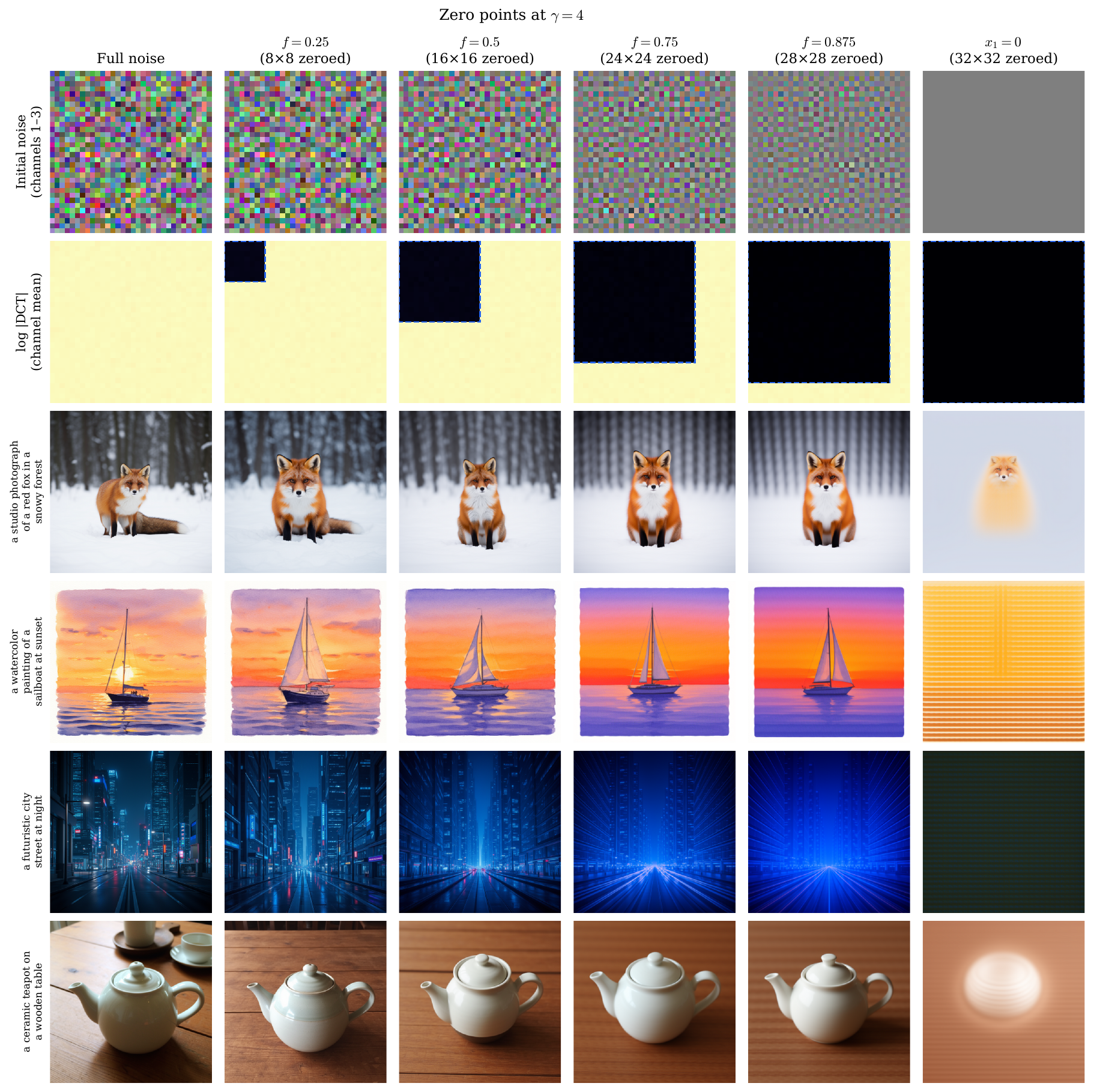}
  \caption{Zeroing increasing fractions $f$ of the low DCT frequencies of a single initial noise $x_1$ for FLUX-Klein-9B at $\gamma=4$. Top: the first three latent channels of the initial state. Second row: the log-magnitude of its DCT, averaged over channels, with the zeroed $k\times k$ block outlined. Remaining rows: the resulting samples for four prompts. As $f$ grows, samples converge to a canonical, centered composition; at $f=1$ ($x_1=0$) they degrade into flat or striped, out-of-distribution images.}
  \label{fig:lowpass-ablation}
\end{figure}

\textbf{Zero-Noise Mode Collapse.}
\label{app:zero-noise-collapse}
For \Cref{fig:zero-noise-collapse,fig:zero-noise-collapse-all} we use FLUX.2-klein-base-9B at $512\times512$ with 20 Euler steps and $\gamma\in\{1,2,4,8,12\}$, sampling 512 initial noises per prompt that are shared across guidance scales. Each prompt names a subject with two plausible modes: the direction a side-on subject faces, or a binary attribute such as color or time of day. A vision-language model (Qwen3.5-9B) answers a multiple-choice question about each image, and we score it by the difference of the log-probabilities of the two answers. For left/right questions we offer a third answer, ``no clear direction,'' and antisymmetrize the score under a horizontal flip of the image, which cancels any labeler bias toward either side. A label counts as confident when the score exceeds 2 in magnitude and, for left/right questions, the third answer has probability at most $0.5$; percentages are over confident labels.
We screened 54 candidate prompts with 32 samples at $\gamma=1$ and kept the 29 with at least 20 confident labels and a minority mode of at least $15\%$. The mode exhibited by $y_{\gamma,0}(0)$ is taken from 8 low-frequency-zeroed samples at $\gamma=12$, and is assigned only when all 8 are labeled confidently and identically; otherwise the prompt has no predicted mode. Intervals are percentile bootstrap intervals over samples (1,000 resamples). Because the initial noises are shared, we test the change between $\gamma=1$ and $\gamma=12$ with an exact McNemar test on the samples whose confident label switches between the two modes. \Cref{fig:zero-noise-collapse} shows the five prompts with the largest increase among those whose $\gamma=1$ share in the predicted mode is below $60\%$.

\begin{figure}[!ht]
  \centering
  \includegraphics[width=\linewidth]{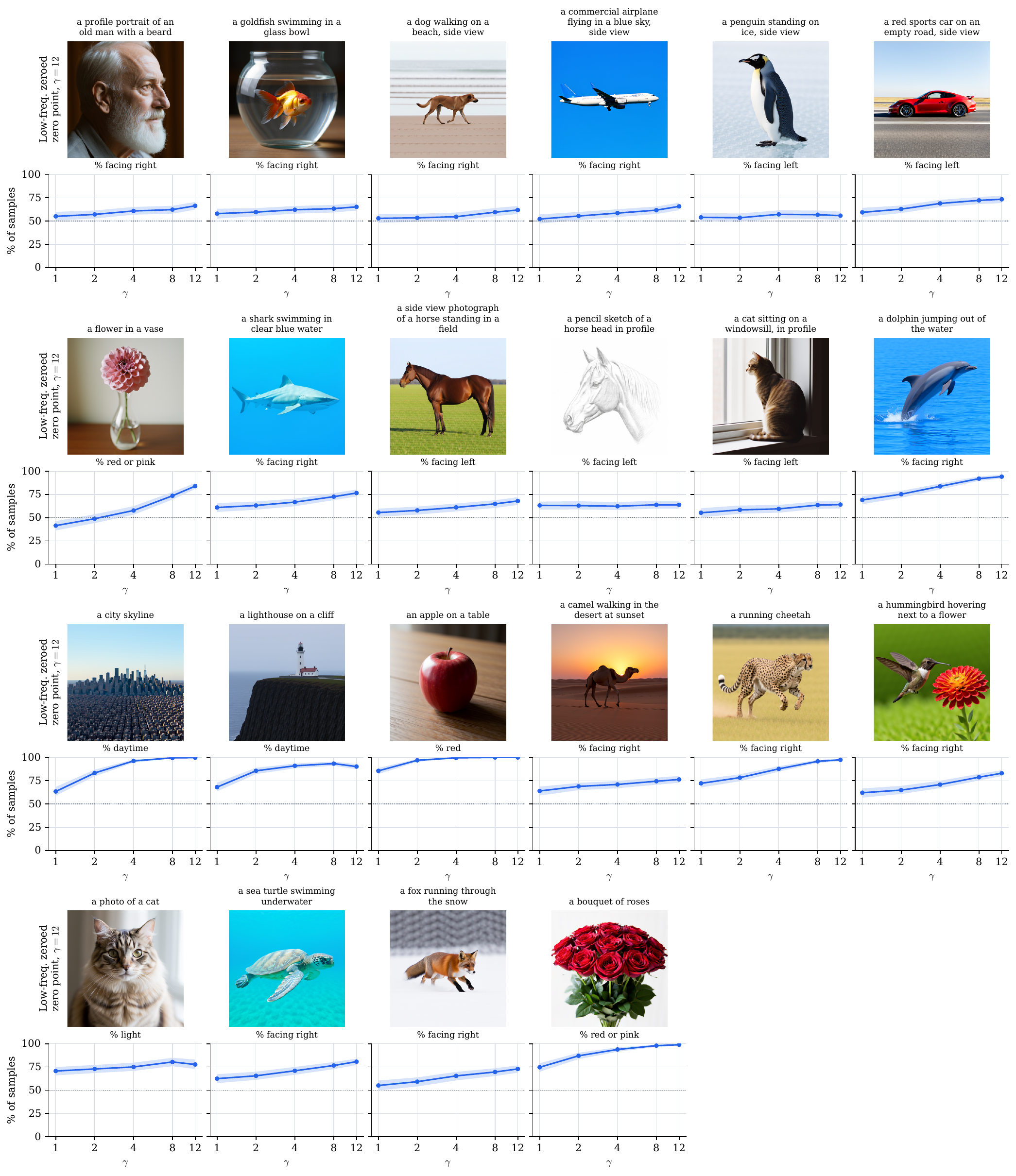}
  \caption{Share of samples in the mode exhibited by the low-frequency-zeroed $y_{\gamma,0}(0)$ at $\gamma=12$ (top of each column), for prompts for which $y_{\gamma,0}(0)$ exhibits a clear mode. Shaded bands are 95\% bootstrap intervals.}
  \label{fig:zero-noise-collapse-all}
\end{figure}

\textbf{t-SNE Embeddings.}
\label{app:image-umap}
For \Cref{fig:imagenet-tsne,fig:image-tsne-sets}, we visualize two-dimensional t-SNE embeddings of DINO-V2 features extracted from generated ImageNet images, with 10 different classes in each plot. We use a diffusion model pretrained with the Flux 2 VAE, with samples generated using DDIM without classifier-free guidance. Conditional samples are generated using class conditioning. Unconditional samples are generated by conditioning on the ``null'' class and selected by rejection sampling with a ResNet-based classifier. The t-SNE plots are each generated in an embedding space using the conditional and unconditional samples corresponding to the 10 selected classes as well as a set of unconditional samples outside of the selected classes. The plots in \Cref{fig:image-tsne-sets} repeat this comparison across different sets of classes.

\textbf{Synthetic Experiments.}
We use the variance-preserving schedule $\alpha(t)=1-t$, $\sigma(t)=\sqrt{1-(1-t)^2}$ and integrate the analytic Gaussian-mixture vector fields from $t=1$ to $t=0$ using fourth-order Runge--Kutta with uniform steps. In \Cref{fig:toy-2d}, the standard setup conditions an unconditional mixture on one component; our setup uses a Gaussian unconditional distribution and a mixture conditional distribution.
The one-dimensional examples use conditional $\tfrac12\Normal(-3,1)+\tfrac12\Normal(3,1)$ and unconditional $\Normal(0,1)$ distributions, with 256 integration steps. \Cref{fig:reweighting-1d} compares exact densities with histograms of 100,000 held-out samples per scale. The synthetic radial panel uses 2.5 million held-out samples per scale. We assign each sample to its nearest component at $\gamma=1$ and project along that component's radial direction, then standardize the projections and plot smoothed histograms.

\begin{figure}[!ht]
  \centering
  \includegraphics[width=0.9\linewidth]{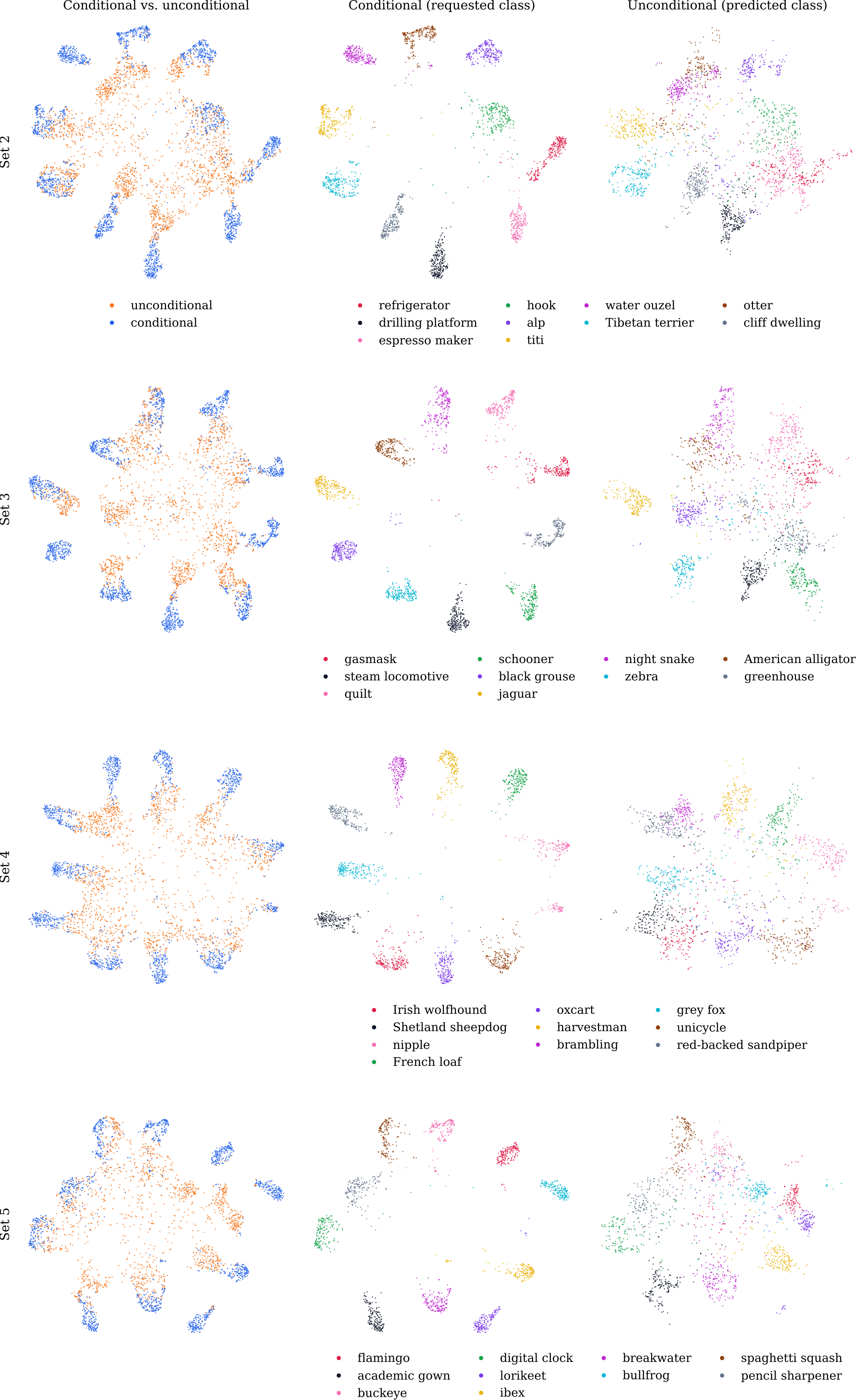}
  \caption{Scatter plots of t-SNE DINO-V2 embeddings for ImageNet conditional and unconditional generations, with 10 different classes in each plot. This shows that the multimodality and complex structure of conditional embeddings relative to unconditional embeddings is robust across different classes. }
  \label{fig:image-tsne-sets}
\end{figure}

%!TEX root = ../main.tex

\clearpage
\section{Sample Galleries}
\label{app:mean-zero-galleries}

\textbf{Sample Galleries.}
We construct mean images by averaging all 10,000 final latents for each condition and scale, then decoding the result. Zero-noise images use the same sampler initialized either at $x_1=0$ (``Zero'') or at a low-frequency-zeroed $x_1$ (``Low-Freq.\ Zero''; \Cref{app:experiment-details}, \Cref{fig:lowpass-ablation}). The galleries show randomly selected samples at $\gamma=1,4,8,12$.

\begin{figure}[!ht]
  \centering
  \includegraphics[width=\linewidth,height=.75\textheight,keepaspectratio]{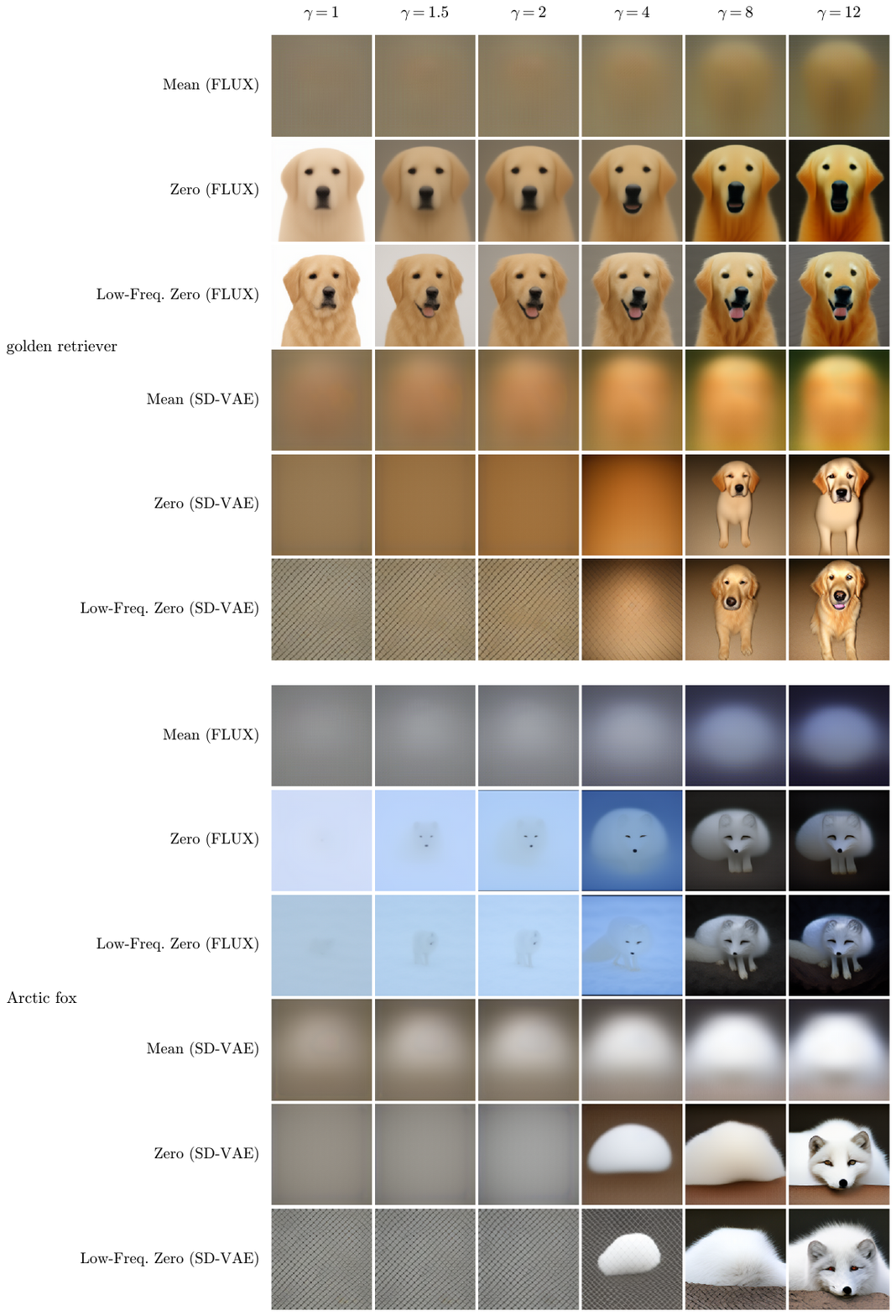}
  \caption{Mean sample, zero-initialized ($x_1=0$), and low-frequency-zeroed $y_{\gamma, 0}(0)$ comparisons for the ImageNet classes golden retriever and Arctic fox (continued in \Cref{fig:stacked-imagenet-classes-cont}). Columns show guidance scales $\gamma$; rows show FLUX and SD-VAE results, as labeled. For FLUX, the low-frequency-zeroed outputs are sharper, more typical versions of the $x_1=0$ outputs. For SD-VAE, both are flat at small $\gamma$ (the low-frequency-zeroed ones with a residual grid texture) and form a recognizable object only for $\gamma\geq4$--$8$.}
  \label{fig:stacked-imagenet-classes}
\end{figure}

\clearpage
\begin{figure}[!ht]
  \centering
  \includegraphics[width=\linewidth,height=.86\textheight,keepaspectratio]{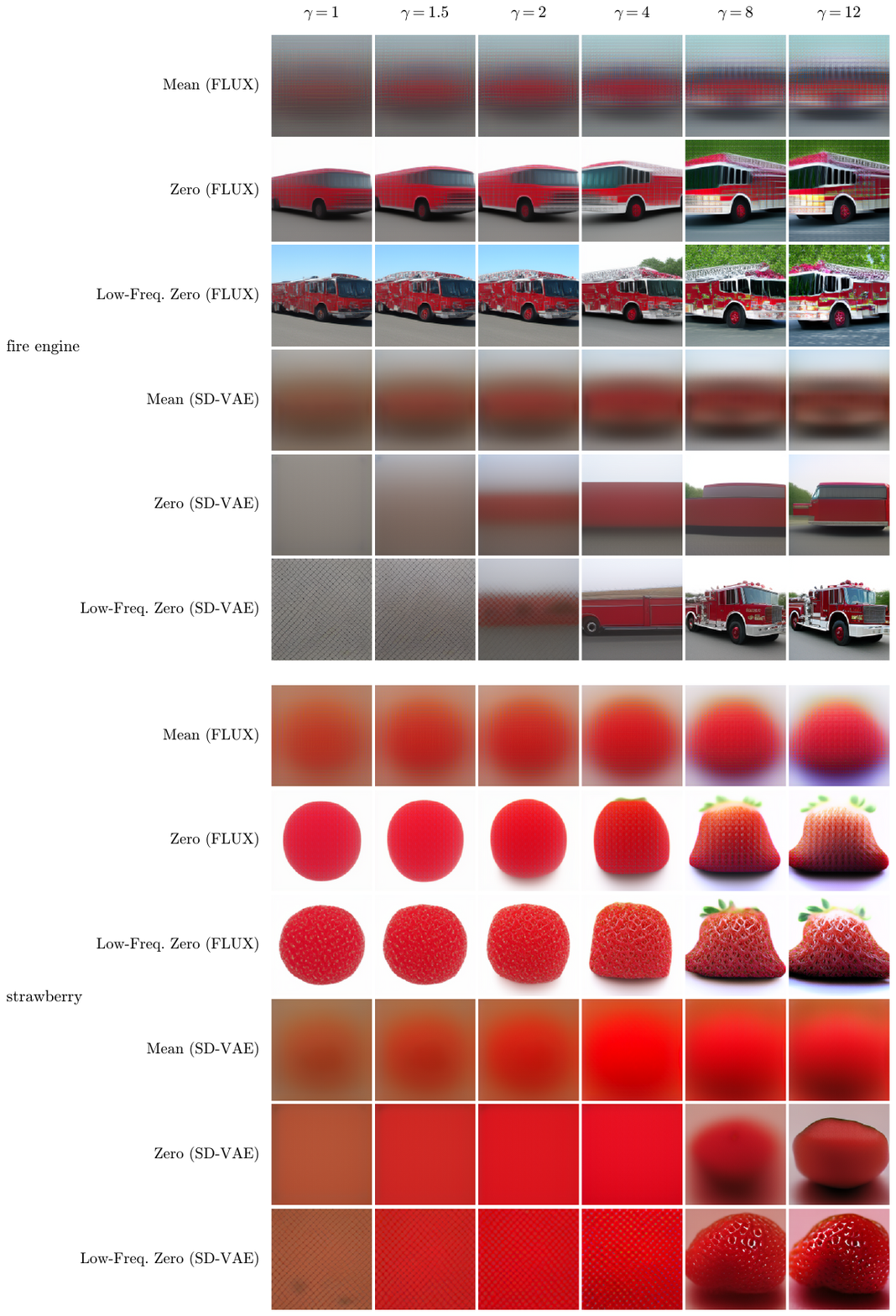}
  \caption{\Cref{fig:stacked-imagenet-classes}, continued: fire engine and strawberry.}
  \label{fig:stacked-imagenet-classes-cont}
\end{figure}

\clearpage
\begin{figure}[!ht]
  \centering
  \includegraphics[width=\linewidth,height=.86\textheight,keepaspectratio]{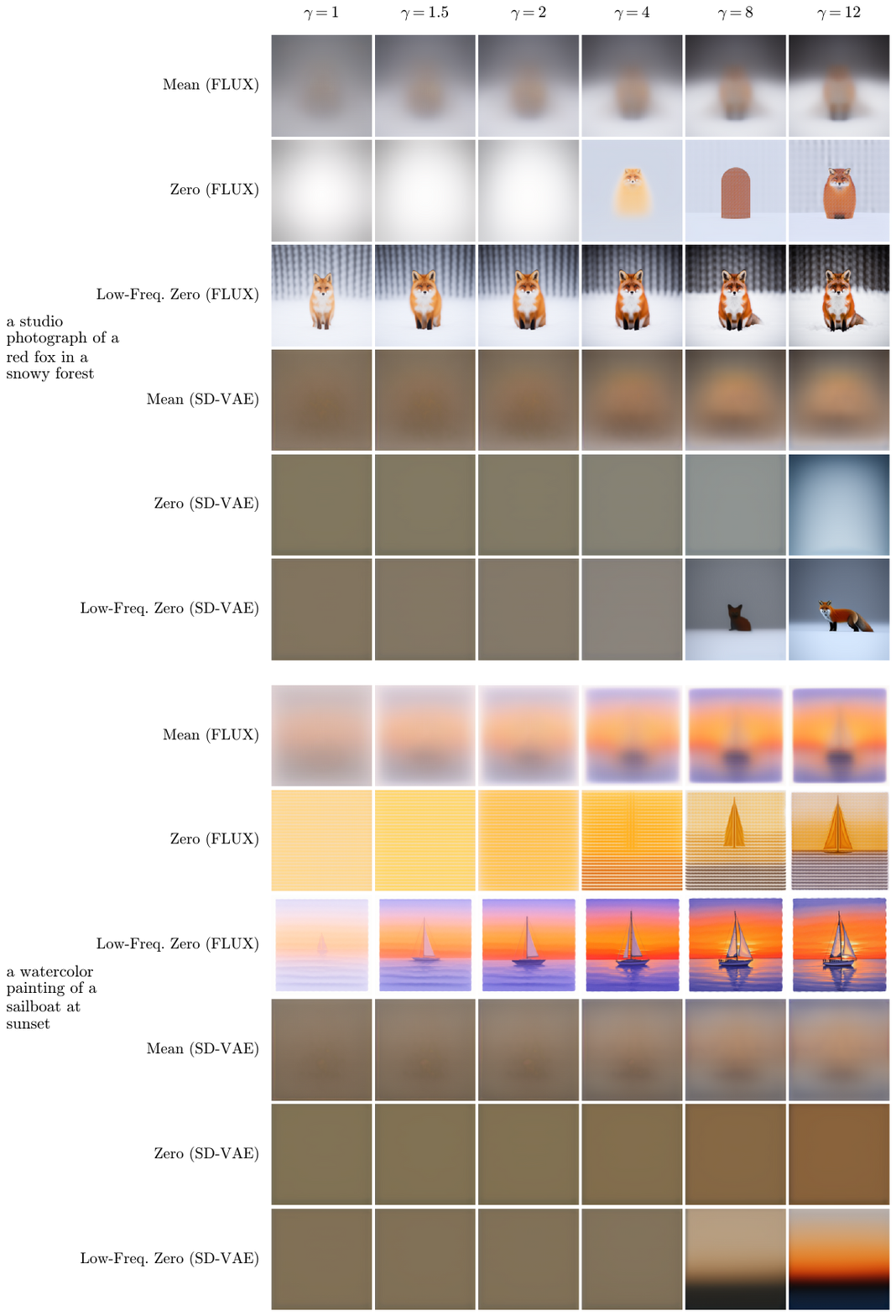}
  \caption{Mean sample, zero-initialized ($x_1=0$), and low-frequency-zeroed samples for the first two text prompts, shown at left (continued in \Cref{fig:stacked-text-prompts-cont}). Columns show guidance scales $\gamma$; rows show FLUX and SD-VAE results, as labeled.}
  \label{fig:stacked-text-prompts}
\end{figure}

\clearpage
\begin{figure}[!ht]
  \centering
  \includegraphics[width=\linewidth,height=.86\textheight,keepaspectratio]{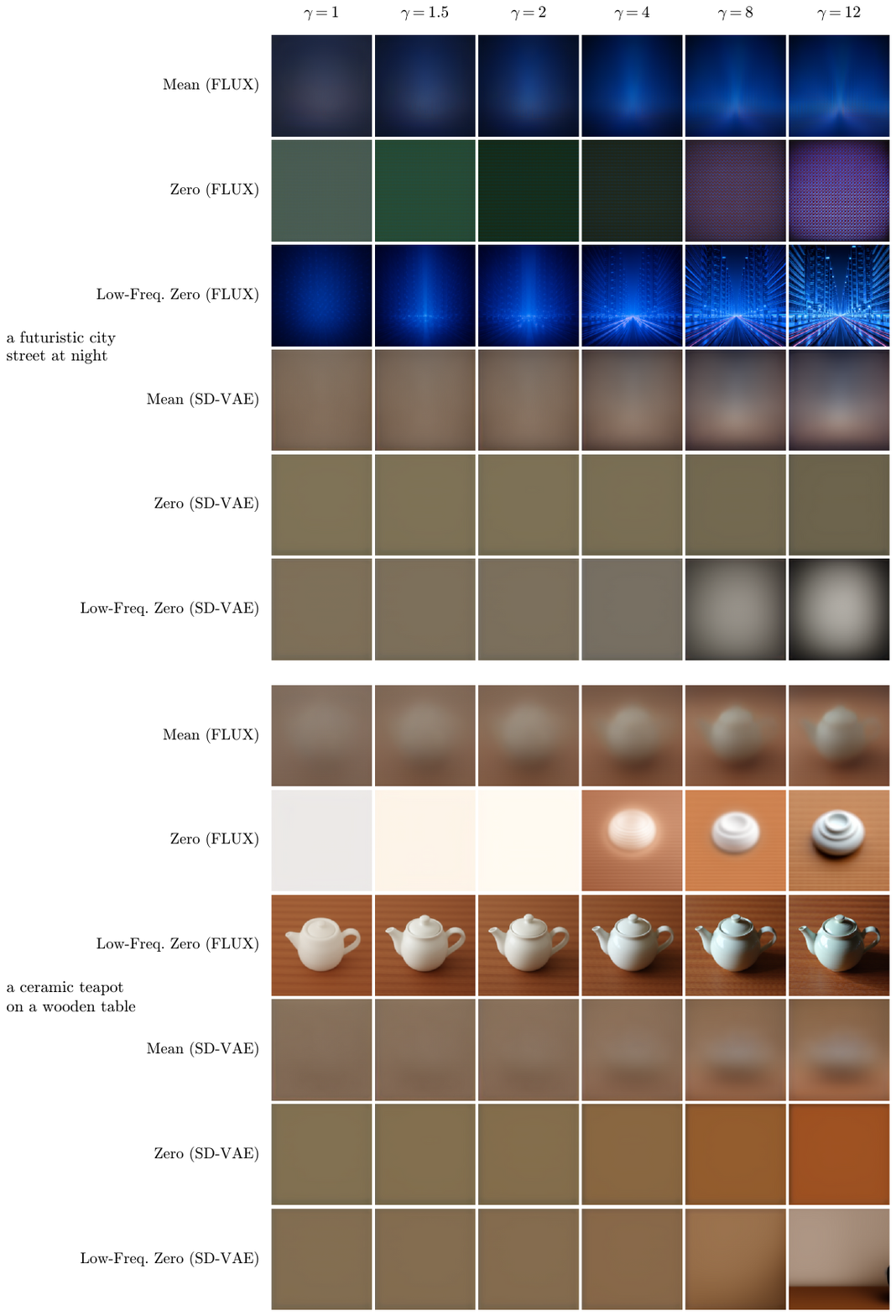}
  \caption{\Cref{fig:stacked-text-prompts}, continued: the remaining two text prompts.}
  \label{fig:stacked-text-prompts-cont}
\end{figure}

\clearpage

%!TEX root = ../main.tex

\label{app:sample-galleries}

\begin{figure}[!ht]
  \centering
  \includegraphics[width=\linewidth,height=.86\textheight,keepaspectratio]{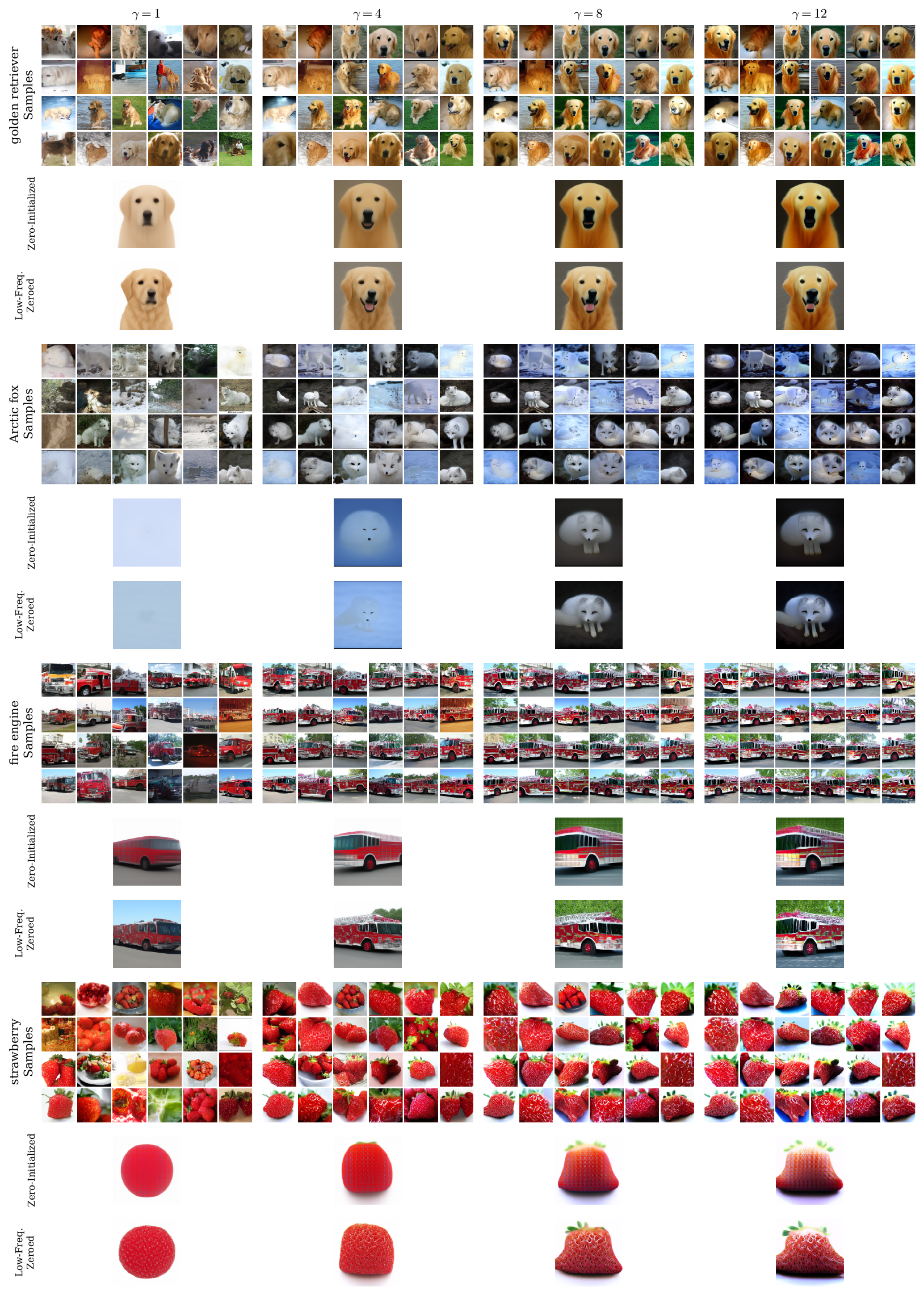}
  \caption{ImageNet FLUX generated samples and zero-noise outputs at guidance
  scales $\gamma=1,4,8,12$ for golden retriever, Arctic fox, fire engine,
  and strawberry. Under each set of samples, rows show the output from
  $x_1=0$ (Zero-Initialized) and from a low-frequency-zeroed $x_1$
  (Low-Freq.\ Zeroed).}
  \label{fig:imagenet-sample-gallery}
\end{figure}

\clearpage
\begin{figure}[!ht]
  \centering
  \includegraphics[width=\linewidth,height=.86\textheight,keepaspectratio]{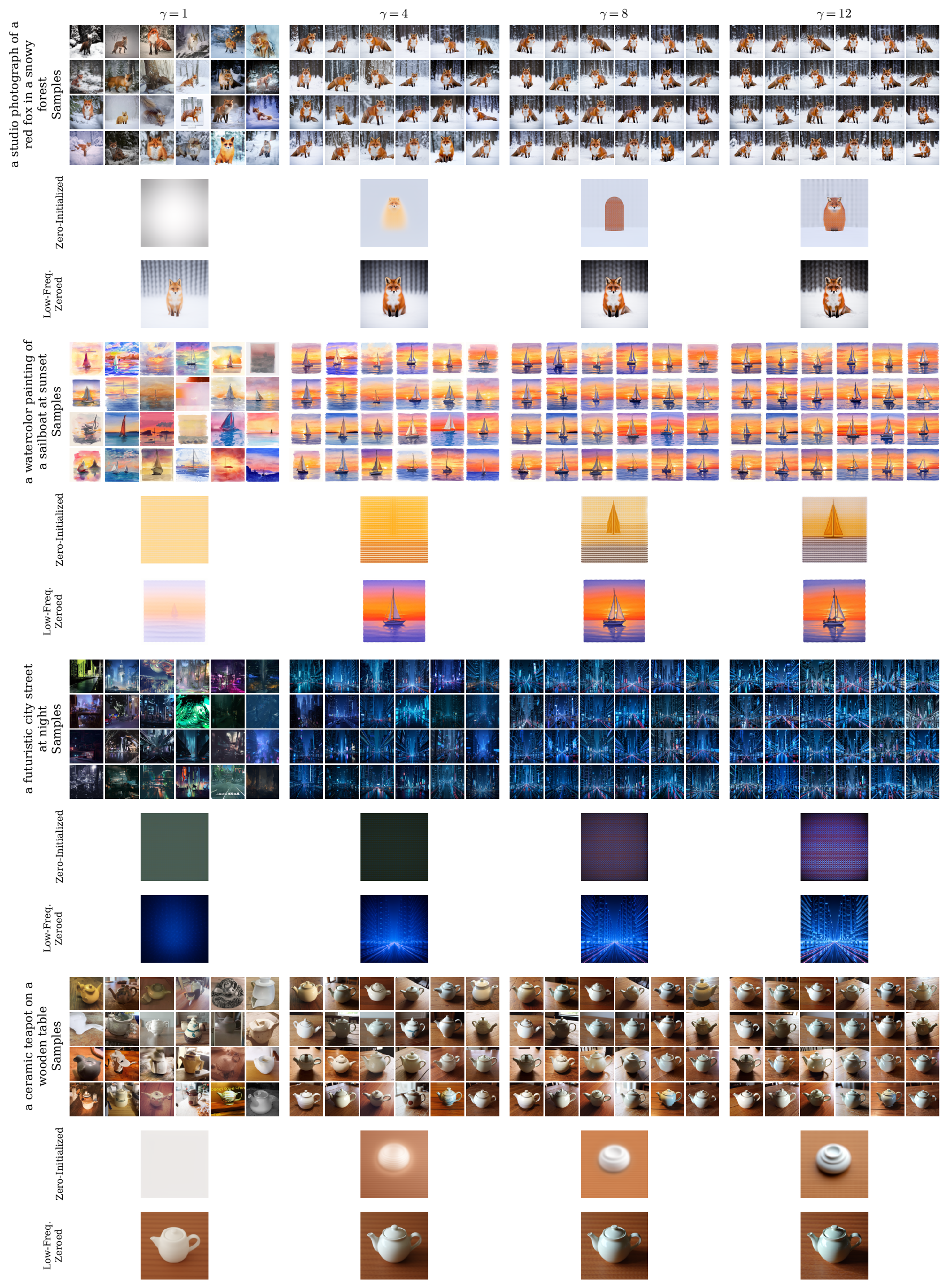}
  \caption{FLUX Klein generated samples and zero-noise outputs at guidance
  scales $\gamma=1,4,8,12$ for the four text prompts indicated at left. Under
  each set of samples, rows show the output from $x_1=0$ (Zero-Initialized)
  and from a low-frequency-zeroed $x_1$ (Low-Freq.\ Zeroed).}
  \label{fig:text-conditioned-sample-gallery}
\end{figure}

\clearpage
\begin{figure}[!ht]
  \centering
  \includegraphics[width=\linewidth,height=.86\textheight,keepaspectratio]{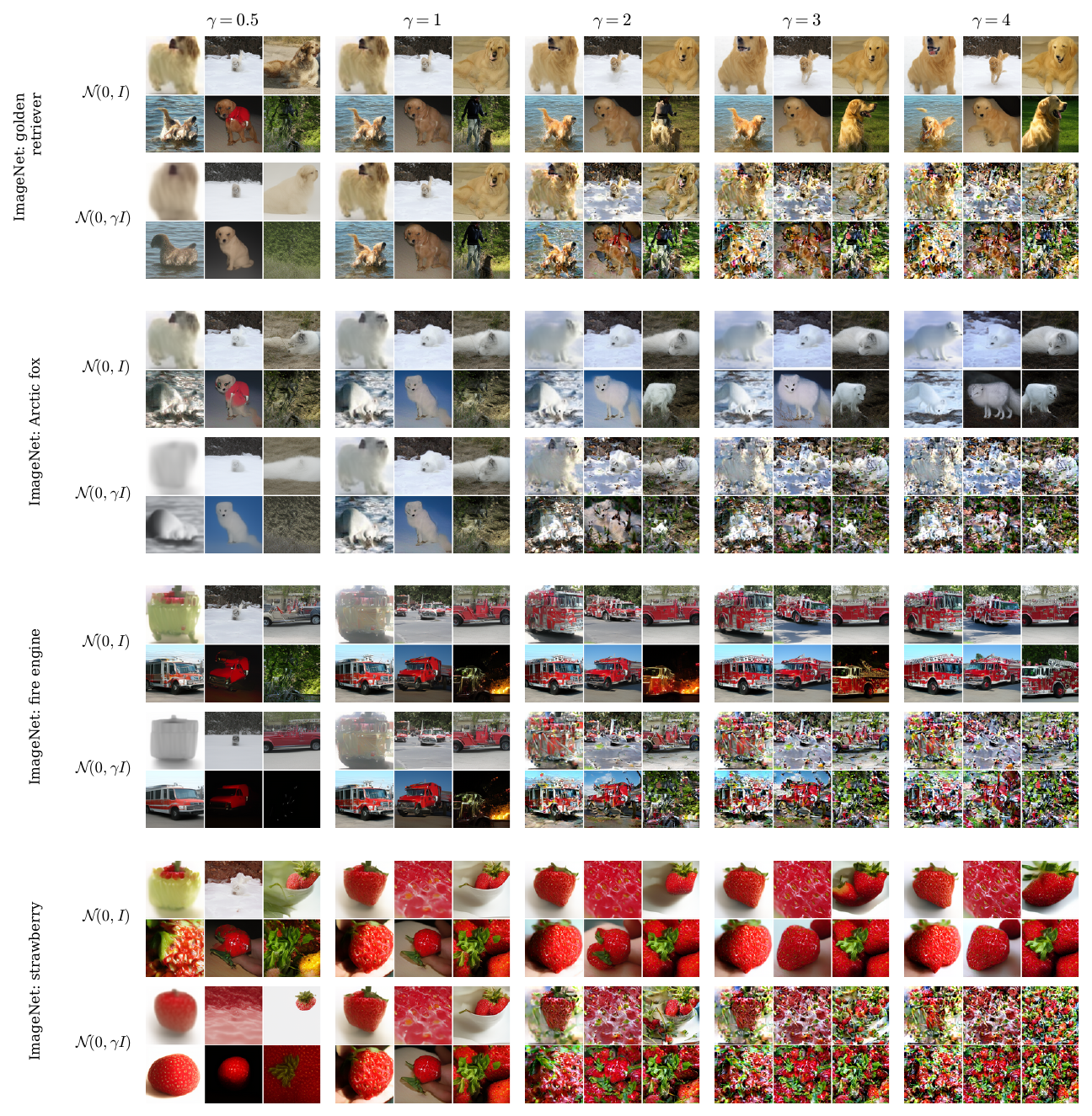}
  \caption{\Cref{fig:noise-scale-main} for the four ImageNet classes with FLUX latents, indicated at left. In each class's group of rows, the top row starts from $x_1=\epsilon\sim\Normal(0,I)$ and the bottom row from $x_1=\sqrt{\gamma}\,\epsilon\sim\Normal(0,\gamma I)$, with the same six $\epsilon$ in each $2\times3$ cell.}
  \label{fig:noise-scale-imagenet}
\end{figure}

\clearpage
\begin{figure}[!ht]
  \centering
  \includegraphics[width=\linewidth,height=.86\textheight,keepaspectratio]{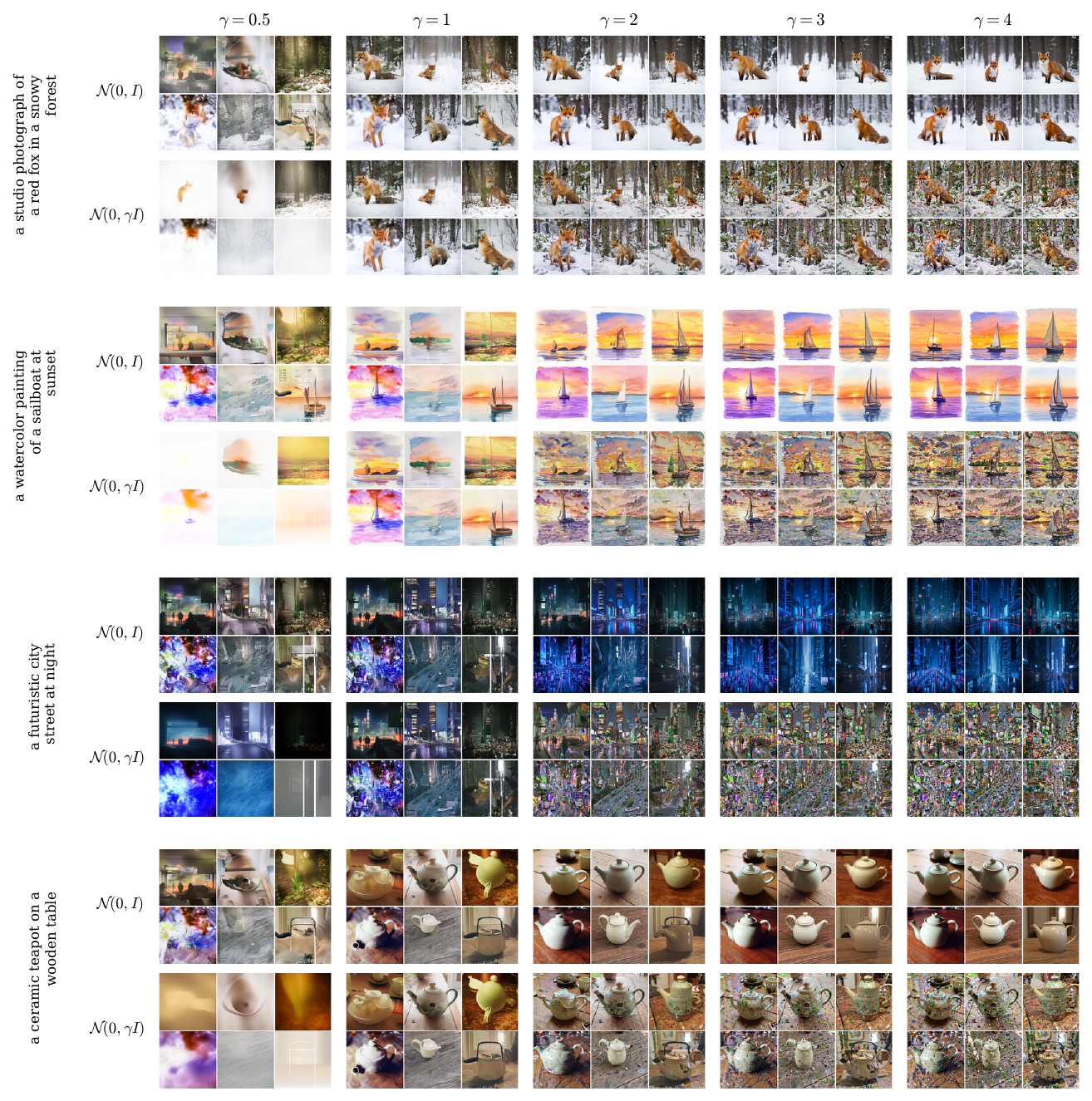}
  \caption{\Cref{fig:noise-scale-main} for all four FLUX Klein text prompts, indicated at left. In each prompt's group of rows, the top row starts from $x_1=\epsilon\sim\Normal(0,I)$ and the bottom row from $x_1=\sqrt{\gamma}\,\epsilon\sim\Normal(0,\gamma I)$, with the same six $\epsilon$ in each $2\times3$ cell.}
  \label{fig:noise-scale-appendix}
\end{figure}

\clearpage

%!TEX root = ../main.tex

\section{Language Model Usage}
\label{app:llm-usage}

Large language models (LLMs) were used for manuscript and proof review, as well as for light editing of the main body and documenting hyperparameters in the experimental details section. Proofs in this manuscript were written and checked by hand, although LLMs were used to check for inconsistencies and issues. LLM-enabled search tools were used for finding some related work, although all cited work was at least in part read manually.

Language models and agentic tooling were heavily used in writing code for the experiments and to create the figures presented in this manuscript. We include all code necessary for reconstructing these figures in the supplementary information.

\end{document}